# Regulating eXplainable Artificial Intelligence (XAI) May Harm Consumers

Behnam Mohammadi[1] NikhilMalik[2] Tim Derdenger[3] Kannan Srinivasan[4]

behnamm@cmu.edu maliknik@usc.edu derdenge@cmu.edu kannans@cmu.edu

[1, 3, 4] Carnegie Mellon University, Tepper School of Business

[2] University of Southern California, Marshall School of Business

**Abstract**

Recent AI algorithms are black box models whose decisions are difficult to interpret. eXplainable AI (XAI) is a class of methods that seek to address lack of AI interpretability and trust by explaining to customers their AI decisions. The common wisdom is that regulating AI by mandating fully transparent XAI leads to greater social welfare. Our paper challenges this notion through a game theoretic model of a policy-maker who maximizes social welfare, firms in a duopoly competition that maximize profits, and heterogenous consumers. The results show that XAI regulation may be redundant. In fact, mandating fully transparent XAI may make firms and consumers worse off. This reveals a tradeoff between maximizing welfare and receiving explainable AI outputs. We extend the existing literature on method and substantive fronts, and we introduce and study the notion of XAI *fairness*, which may be impossible to guarantee even under mandatory XAI. Finally, the regulatory and managerial implications of our results for policy-makers and businesses are discussed, respectively.

## 1. Introduction

Recent years have seen a surge in the adoption of Artificial Intelligence (AI) models for decision-making. Gartner identifies AI engineering among the top 12 strategic technology trends of 2022[1,a], and International Data Corporation (IDC) forecasts global spending on AI systems will jump from \$85.3 billion in 2021 to more than \$204 billion in 2025[2]. In marketing, various forms of AI models have been used over the years. Segmentation, for instance, benefits from AI in order to obtain tourist segments based on the meaning of destinations (Valls et al., 2018), micro-segment retail customers based on their preference for personalized recommendation (Dekimpe, 2020), and run psychographic consumer segmentation in the art market (Pitt et al., 2020). In targeting, AI has been used to profile digital consumers using online browsing data (Neumann et al., 2019), target important customers by estimating their hazard function (Drew et al., 2001), and identify the best target for proactive churn programs (Ascarza, 2018), to name a few. AI could also help brands develop successful positioning statements and compelling slogans (Huang and Rust, 2021).

[a] Footnotes are numbered as a, b, c, … while endnotes are numbered as 1, 2, 3, … .

That being said, a key challenge in adoption of AI is the interpretability of its decisions or predictions. While early AI models were easily interpretable, the latest methods such as Deep Neural Networks (DNNs) are opaque decision systems with gigantic parametric spaces that make their decisions difficult to understand even by their creators[3,a]. In this regard, most recent AI algorithms are complex *black box* models (Castelvecchi, 2016). Sometimes, this is not a problem because the consequences may be negligible (e.g., email categorization AI), or because users may not want to know explanations for AI outputs (e.g., answers given by a voice assistant like Siri or Alexa). But in most settings, humans are generally reluctant to adopt algorithms that are not interpretable, tractable, and trustworthy (Zhu et al., 2018), especially given multiple incidents where black boxes resulted in biased outcomes. Google, for example, showed fewer ads related to high-paying jobs to women than to men (Datta et al., 2015), Amazon's same-day delivery bypassed black neighborhoods[4], and the software on several types of digital cameras struggled to recognize non-white users' faces[5,6].

To address AI interpretability, researchers have recently shifted their focus to eXplainable AI (XAI), a class of methods that aim to produce "glass box" models that are explainable to humans while maintaining a high level of prediction accuracy[7] (Abdollahi and Nasraoui, 2016; Csiszár et al., 2020; Doshi-Velez and Kim, 2017; Holzinger et al., 2017b; Lipton, 2017; Murdoch et al., 2019). Put differently, the goal of XAI is to enable human users—including non-technical non-experts—to understand, trust, and effectively manage the emerging AI systems. XAI has been gaining traction across healthcare, retail, media and entertainment, and aerospace and defense. According to Explainable AI Market Report[b], the market size of XAI across the globe was estimated to be $4.4 billion in 2021 and is predicted to reach $21.0 billion by 2030 with a compound annual growth rate (CAGR) of 18.4% from 2022 to 2030. But despite the growing interest in XAI—from researchers to engineers and governments[8]—little is known about the *economic implications* of XAI for firms and consumers (Adadi and Berrada, 2018). Consumer activists typically favor regulating AI in general and mandating increasingly transparent XAI in particular[9]. Our paper provides a timely study of the economics of AI and XAI for firms and consumers.

We model a duopoly market where firms offer a product based on AI algorithms, e.g., lending based on an AI decision-maker. Such AI algorithms have been used by marketers to slice loan borrowers into those who will pay back the loan and those who will not (Netzer et al., 2019). We let firms set product quality and price while consumers choose to purchase the product from one of the two firms. AI product quality can be related to the accuracy of the AI model or unrelated to the AI model such as timely disbursement of loan. The policy-maker aims to maximize total welfare by choosing the XAI level ranging from no explanation to full explanations. In the loan example at full explanations, the AI-based loan rejection decision could be fully explained to the applicant by breaking down how various characteristics of their loan application (e.g., credit history, employment, purpose of loan) contribute to their final score. Customers are heterogenous in their preferences for quality and explanations, and firms engage in a three-stage game to capture market share by choosing their explanation strategy, quality, and price levels, respectively. In our model, firms are horizontally differentiated in XAI method and vertically differentiated in quality.

[a] For example, OpenAI's GPT-3 model has about 175 billion parameters (Brown et al., 2020).

[b] By Next Move Strategy Consulting, nextmsc.com

While most current regulations do not mandate where and how firms can use black boxes[10], the de facto stance involves *mandating* businesses using automated decision-making systems (e.g., AI models) to explain their decisions to end-users (Pradhan et al., 2022). We incorporate this into our model by considering a mandatory XAI regulation where the policy-maker chooses the XAI level and firms must offer XAI at this level. In addition, we contribute to the XAI policy-making debate by introducing a new regulatory lever of *optional* XAI where the policy-maker chooses the XAI level, but firms are free to decide whether or not they will offer XAI at this level. One of our findings confirms the popular belief that mandating XAI often makes the society better off. But we also identify conditions where **mandatory XAI may have no additional benefit over optional XAI**. While we do not explicitly model the regulatory cost of monitoring compliance, this finding can lead to significant savings in regulatory oversight while delivering nearly similar outcomes. As we will discuss, the reason for this counterintuitive result is that optional XAI gives firms more freedom, allowing them to differentiate on XAI to avoid unsustainable price wars and even provide more differentiated products for their heterogeneous consumer demand, which in turn could increase the welfare.

Closely related to calls for mandatory XAI is the call for full transparency to consumers (Liu and Wei, 2021; Patel et al., 2022; Pollack, 2016). The general belief is that full-explanations are a win-win for firms and consumers if XAI implementation costs, model privacy, and manipulation concerns are disregarded. Surprisingly, we find that under both mandatory and optional XAI, **requiring full-explanations may actually make firms and consumers worse off**. This result—that full-explanations are not necessarily the optimal policy-making strategy—is robust even when the total welfare objective is swapped for other objectives that are more convenient to measure (and publicize) such as total number of firms that offer XAI, average XAI level received by consumers, and XAI fairness (discussed below). Moreover, conventional wisdom would suggest that under optional XAI—where firms have more freedom—the policy-maker needs to set a more conservative, low XAI level to incentivize firms to offer XAI.

But we show that the optimal choice for the policy-maker may be to set a *higher* XAI level under optional XAI.

A natural follow-up question is: What would firms do if left unregulated? Firms might currently point to several reasons for not offering full XAI, such as long-term implications of stolen IP, copycats, adversarial attacks, or unavailability of XAI methods applicable to their AI model (Cinà et al., 2022). While some papers examine the interrelation between XAI and adversarial attacks (Galli et al., 2021; Kuppa and Le-Khac, 2020), to the best of our knowledge, no studies have documented evidence that XAI has facilitated or enabled such attack vectors. Therefore, we consciously choose not to model adversarial threats. In addition, nowadays there exist cutting-edge XAI methods that are provided as free and open-source packages. Thus, we set the cost of adopting XAI methods to zero. At the same time, we do recognize the challenges in organization adoption of AI and XAI strategies, and we build that friction into our model. Under a laissez-faire attitude policy, the results indicate that an **unregulated market can be as good as regulated** (or even better in settings where firms are limited in price and quality choices) in terms of total welfare, total consumer utility, and average XAI level offered to consumers. The direct impact of XAI is to provide greater utility for consumers and therefore room for all firms to potentially extract some of the additional surplus by increasing prices, which explains why optional regulation or unregulated markets work well. Thus, calls for instituting costly mandatory regulations may be redundant.

Firms face a growing pressure by legislators and customers alike to adhere to accountable AI practices such as algorithmic transparency and XAI (Ukanwa et al., 2022). Yet, little is known about the economic consequences of adopting XAI for firms. Our paper sheds light on this problem by offering two unexpected insights for managers. First, we point out the **significance of quality as the prerequisite product attribute for developing profitable XAI strategy**. This holds even when the market values explanations more than quality. Second, in the absence of regulations, we show that **firms' best strategy may be to *mirror* each other's XAI level while using different XAI methods**. This surprising result is added motivation why we study regulations where the same XAI level is prescribed or mandated for both firms.

Finally, we contribute to the AI fairness literature by introducing and investigating the notion of *XAI fairness*. This is a timely treatment of the subject because XAI is often touted as an enabler of AI fairness by detecting algorithmic bias (Zhou et al., 2022), and yet it is unclear how different market segments (protected and unprotected groups) can benefit from XAI. Our paper addresses this question and proves that **it is impossible to guarantee XAI fairness**. We show that a fundamental requirement for fairness is *symmetrical* XAI levels of the firms, and while mandatory XAI—by design—would ensure this symmetry, there are cases where firms under optional XAI or even unregulated firms choose symmetric XAI levels. Our paper then discusses the optimal policy-making strategy if the objective is XAI fairness.

To our knowledge, this is the first attempt towards systematically and simultaneously representing AI products (price and quality, e.g., through accuracy), XAI (level, method, and fairness), and heterogeneous customer preferences for quality and XAI. Our model is generalizable across AI algorithms and XAI methods. It also encompasses multiple regulatory levers (mandatory, optional, and no regulation) as well as multiple policy objectives (total welfare, consumer welfare, average explanations, and XAI fairness). Obtaining tractable closed-form solutions for such a comprehensive model is challenging. This is achieved only by breaking down strategy spaces into regions (e.g., firm competition dominated by quality vs. explanation) where welfare and profit functions are analytically solvable. The thoroughness of the model is not an arbitrary engineering goal, but a conscious decision to model all inter-related economic forces (price, quality, and explanations) while assuming away the non-economic factors (e.g., stolen IP and adversarial attacks). This is unique among the literature in this area and hopefully provides a framework for more research questions to be answered.

## 2. Background

XAI methods often use state-of-the-art human-computer interfaces (mostly visual), such as Figure 1. Most of the XAI scholarship focuses on local explanations that justify a specific decision or prediction[a]. Local explanations are generated by methods such as LIME, LOCO, and SHAP (Lei et al., 2017; Lundberg and Lee, 2017; Ribeiro et al., 2016). But not all local explanations are equally local: XAI methods can be tuned to reveal different amount of information in the explanations. For example, Facebook provides some local explanation for why a certain ad was shown (Figure 1). Facebook's ad targeting AI likely uses dozens of customer features, but in this specific example they explain only three features that resulted in the customer receiving a particular ad. In comparison, Google Search page informs the user that a certain ad was shown because of "your current search terms". This provides only a global description of how the AI works and offers zero local explanation[b].

[a] See Appendix 0 for local vs. global explanations and more visual examples of local XAI.

[b] The more local an explanation is, the more personalized it becomes. Conversely, global explanations are not tailored to each consumer and may apply to several people.

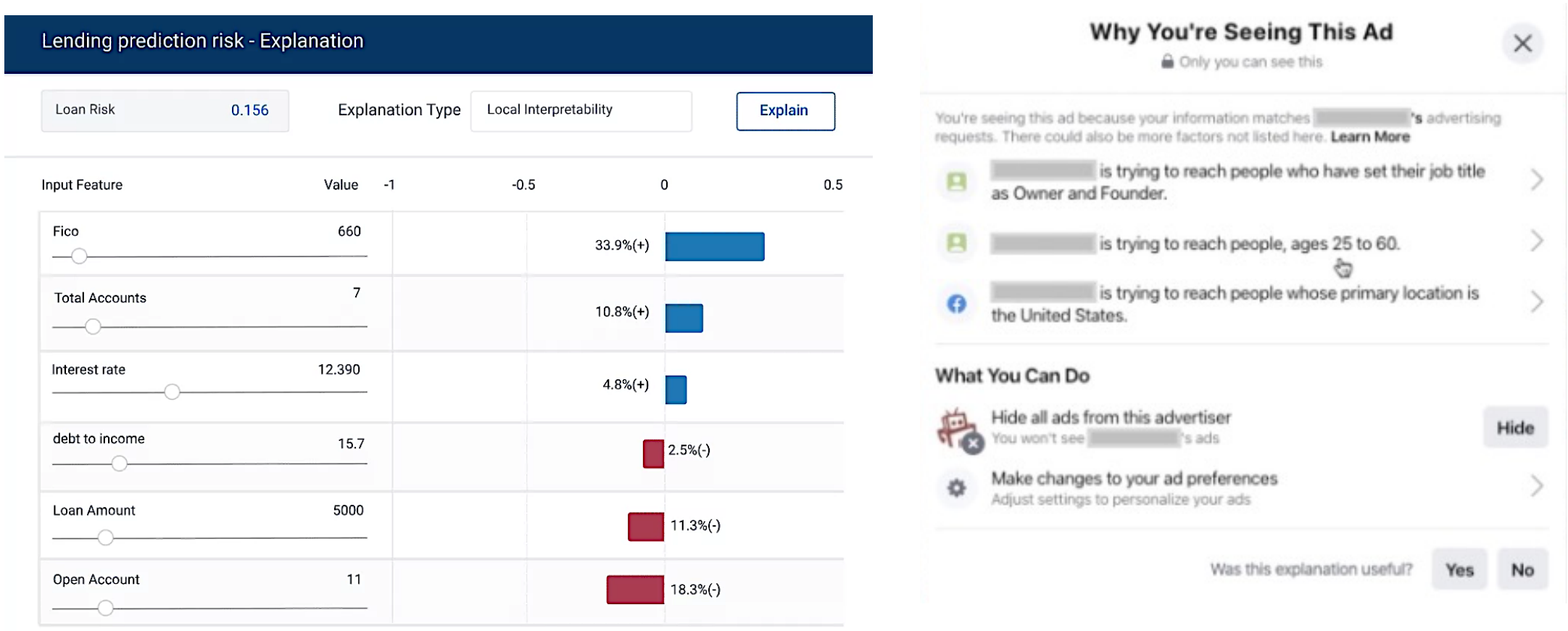

Figure 1. (Left) When applying for a loan, XAI analyzes each feature and finds its positive/negative effect on the outcome (loan risk). Source: AIMultiple[11]. (Right) Facebook provides some explanation about why it shows an ad to you.

## 2.1. The Need for XAI

The term "XAI" was coined in 2004 (van Lent et al., 2004), but the problem of explainability has existed since the work on expert systems in the mid-1970s (Moore and Swartout, 1988). It was only after the proliferation of AI across industries in recent years that social, ethical, and legal pressures began calling for new AI methods that are capable of making decisions explainable and understandable[12] (Adadi and Berrada, 2018). In regard to social justice and fairness, it is well-documented that AI systems might yield biased results (Caruana et al., 2015; Howard et al., 2017). For example, COMPAS (Lightbourne, 2017, Tan et al., 2018), a black-box recidivism risk assessment software has been criticized for violating the due process rights because it uses gender and race to predict the risk of recidivism. If Judges are to trust the decisions made by such AI systems, some transparency and information is required to ensure that there is an auditable way to prove that decisions made by the algorithm are fair and ethical.

Another area where fair decisions are of highest importance and could benefit from XAI is financial services. In the US, the Fair Credit Reporting Act (FCRA) requires consumer reporting agencies to provide a list of the key factors that negatively influenced the consumer's score[13] (McEneney and Kaufmann, 2005). Consequently, loan issuers are required by law to make fair decisions in their credit score models and provide the needed "reason code" to borrowers who were denied credit. This has pushed some credit bureaus such as Experian and Equifax to work on AI-based models that generate more explainable and auditor-friendly reason codes[14].

XAI is also finding its place in some emerging fields such as autonomous vehicles (Bojarski et al., 2016; Haspiel et al., 2018). In fact, a number of recent accidents (some fatal) have fueled societal concerns about the safety of this technology[15] (Stanton et al., 2019; Yurtsever et al., 2020). Thus, the real-time decisions of autonomous vehicles need to be explainable in the sense that they are intelligible by road-users. Current and next-generation intelligent transportation systems must comply with the set of safety standards established using the right of explanation[16] ("Dentons," 2021).

Finally, recent research in medical diagnosis has focused on making clinical AI-based systems explainable (Ahmad et al., 2018; Caruana et al., 2015; Che et al., 2016; Holzinger et al., 2017a; Katuwal and Chen, 2016). The urgency of XAI in this domain is not just due to trust concerns—it can literally save lives. For example, in the mid-1990s, researchers trained an artificial neural network (ANN) to predict which pneumonia patients should be admitted to hospitals and which should be treated as outpatients. But the AI would not admit pneumonia patients with asthma on the grounds that it thought they have a lower risk of dying, which is both medically incorrect and dangerous (Adadi and Berrada, 2018). It turned out that the reason was that in the training dataset, pneumonia patients with asthma would often get admitted not just to the hospital but sent directly to the ICU and treated intensively. Since most would survive, the AI concluded that pneumonia and asthma together do not increase the risk of death. Only by interpreting the model using XAI methods can we discover such life-threatening issues and prevent them.

## 2.2. XAI and AI Regulations

The growing body of research on XAI coincides with the emerging issue of the regulatory and policy landscape for AI in jurisdictions across the world (Law Library of Congress (U.S.)., 2019). In 2016, the EU introduced a *right to explanation* of algorithmic decisions in General Data Protection Right (GDPR)[a] which gives individuals the right to request an explanation for decisions made by automated decision-making systems that affect the individual significantly, especially legally or financially (Goodman and Flaxman, 2017). GDPR has garnered support both from consumer right groups such as The European Consumer Organization[17] and from firms such as Facebook/Meta who view the law as an opportunity to improve their data management[18,19,20]. There have been calls for GDPR-style laws to be adopted in the US as well[21,22]. FCRA is the closest equivalent of GDPR in the US as of now. Moreover, France, Germany, and the UK operate under a *comply-or-explain* model of corporate governance[23] which obligates private corporations to adhere to a corporate governance code. Should they depart from the code in any way, they must publicly explain their reasons for doing so (Doshi-Velez et al., 2017). In addition, France has amended its administrative code with the Digital Republic Act which creates a right for subjects of algorithmic decision-making by public entities to receive an explanation of the parameters (and their weighting) used in the decision-making process[24].

However, some critics view AI regulations such as GDPR as unnecessary or even harmful. They argue that regulating AI might stifle many of its social and economic benefits and restrain innovation[25]. Others, such as Intel's CEO Brian Krzanich, argue that it is too early to regulate AI as it is still in its "infancy"[26]. Critics of GDPR, in particular, highlight the fact that many AI algorithms are not intrinsically explainable. While XAI seeks to alleviate this problem (Miller, 2019; Mittelstadt et al., 2019), some critics have called into question the value of XAI as well. Among them is Peter Norvig, former Google research director who argues that humans are not very good at explaining their decisions either, and that the credibility of the outputs of an AI system could be better evaluated by observing its outputs over time (Adadi and Berrada, 2018). If human decisions can depend on intuition or a "gut feeling" that can hardly be put into words, it can be argued that machines should not be expected to meet a higher standard.

[a] See Articles 13 through 15: https://tinyurl.com/EU-regulation-2016-679

## 2.3. Literature

This paper is at the intersection of three growing streams of research. First, an area of research looks at potentially negative implications of unexplained AI on society (Fu et al., 2022; Malik, 2020) and consumers' negative perception from unexplained AI decision changes (Bertini and Koenigsberg, 2021). Another field of research examines concerns with explainable AI arising from tradeoff with AI accuracy (Adadi and Berrada, 2018), privacy of firms' proprietary secrets (Adadi and Berrada, 2018), potential for adversarial attacks by competitors (Goodfellow et al., 2015), and strategic manipulation by consumers (Wang et al., 2022). The third stream of research examines AI algorithms deployed by competing firms. For example (Assad et al., 2020) study endogenous collusive learning in the AI algorithm. In the next section we lay out our model and contrast that with some of this related literature.

# 3. Theoretical Model

We study a duopoly where two firms, indexed $i = 1, 2$, are vertically differentiated in product quality and horizontally differentiated in XAI method. Each firm markets a product whose quality $q_i \geq 0$ and price $p_i \geq 0$ are observed by customers. Without loss of generality (WOLOG), we assume that firm 1 is the high-quality firm and firm 2 the low quality one, that is, $q_1 \geq q_2$. A quadratic cost is considered for quality as $\beta q_i^2$.

We model customer preferences using a characteristics-based approach, defined directly over the attribute dimensions of the available products. Customers are heterogeneous in terms of their valuation of quality (i.e., willingness to pay). This heterogeneity can be motivated, for example, by difference in income levels and is captured by parameter $\theta$ which is uniformly distributed over the population. WOLOG, $\theta$ is normalized to $[0, 1]$. Customer $j$'s flow utility derived from quality of firm $i$ is $u_{ij}^q := \theta_j q_i$ where $\theta_j$ is the customer's sensitivity to quality. Everyone has the same preference for price so that those who purchase from firm $i$ experience the same price flow (dis)utility $u_{ij}^p := -p_i$. The third component of firms' products is the explanation that they offer as the reasoning behind the AI decision. The explanation has two characteristics: XAI level $\xi$ and XAI method $e$ as follows:

**XAI level $\boldsymbol{\xi}$** represents the amount of information contained in the explanation. Consider a loan approval AI that models applicant scores as a function of 8 application features (e.g., credit history, employment, purpose of loan). A high XAI level would explain a loan rejection decision to the applicant by breaking down how, say, 6 out of 8 characteristics of their loan application contribute to their final score. A low XAI level would only explain, e.g., 2 out of 8 characteristics. Instead of number of features explained, one could alternatively interpret XAI level as R-square explained. In an extreme case, zero XAI ($\xi = 0$) would be akin to no explanation.

In our model, customers prefer higher XAI level. But AI outputs for some customers may require more information to be explained than other customers. Consider again the loan application scoring AI using a decision tree model (Figure 2). The decision tree splits the entire sample iteratively using informative features. Eventually, each branch in the tree terminates because the samples remaining in the terminating node cannot be further split into more informative nodes. Each customer prediction is in exactly one of these terminating nodes. One way to explain the prediction would be to present to the customer all the splits from the starting node to the terminating node and how each split contributed to the customers' score. Since some terminating nodes have lower depth, they can be explained by explaining fewer splits. The XAI level $\xi$ here would refer to the number of splits or features explained. If all splits are explained ($\xi = 1$), then all customer predictions get explained irrespective of their depth. If only some splits are explained (e.g., $\xi = 0.5$), this may fully explain some predictions at a shallow depth but not others. The same intuition can be derived from a one-dimensional example where some representative points are explained in detail (Figure 3). If a customer's characteristics happen to be at or very close to one of the representative points, they will get a satisfactory explanation for themselves. As XAI level increases, more representative points are explained, and consequently more customers are satisfied by the explanations.

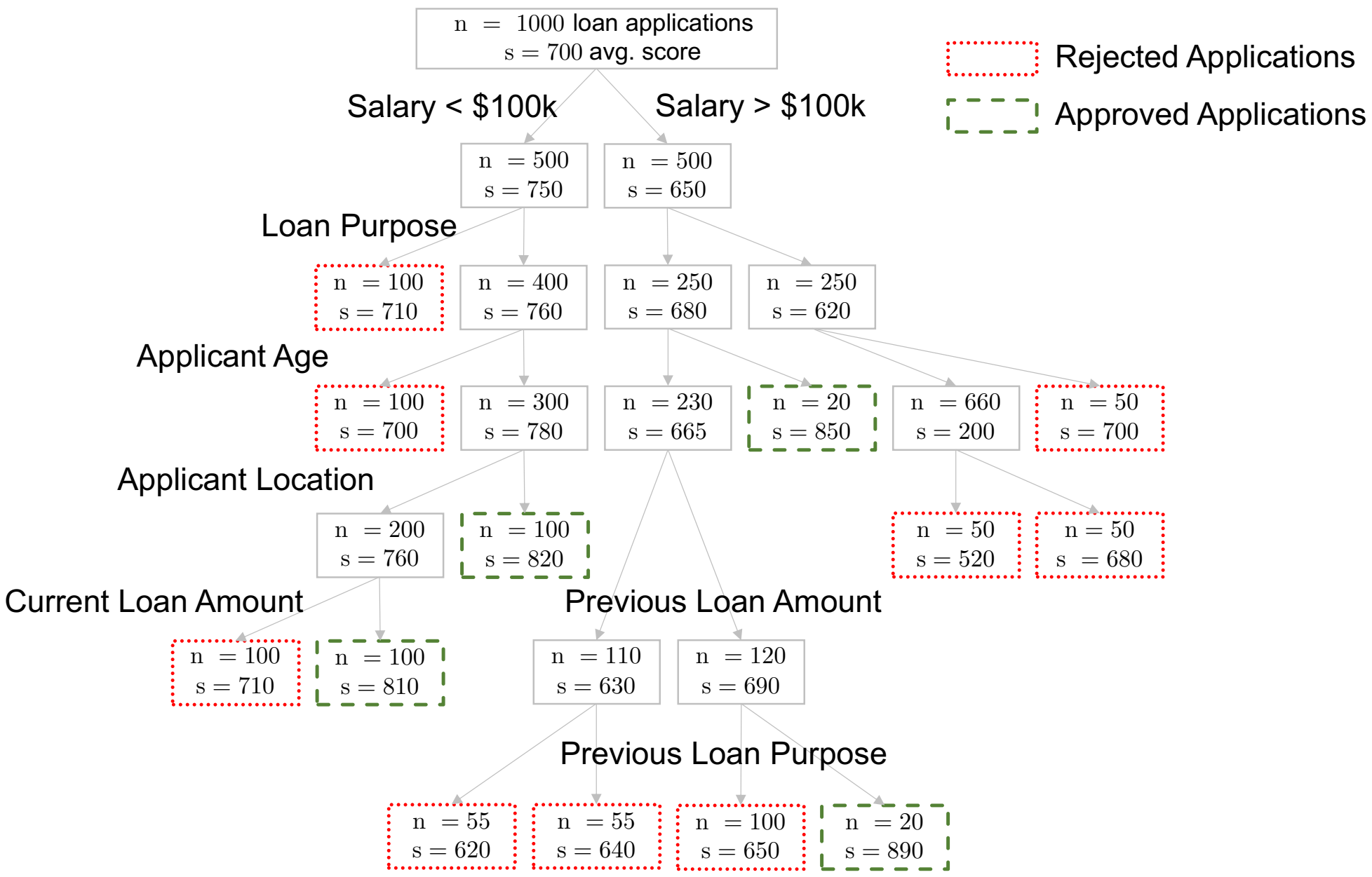

Figure 2. A decision tree that scores loan applications and approves the application if the predicted score is > 800. The tree branches terminate when a node has $\leq 100$ samples. Rejected applications can be explained based on all tree splits. Some rejected applications can be explained using two features, others can be fully explained using 5-6 features.

To model the relationship between customer's flow utility and explanations, we consider a Hotelling line such that customer AI predictions are uniformly distributed on the $x$ axis over $[0, 1]$. We can think of firm $i$'s XAI as located at a single point on this line at $e_i$. At a low XAI level, the firm explains well very few predictions close to $e_i$. Customer $j$'s flow (dis)utility due to lack of explanations from firm $i$ is denoted by $u^e_{ij}$ and is increasing in the distance between the customer's preference $x_j$ and firm $i$'s location $e_i$ on the Hotelling line, that is, $u^e_{ij} := -t|x_j - e_i|$ where $t > 0$ is the "transportation cost" in the Hotelling model and can be thought of as the per unit cost of misfit. An increasing XAI level can be modelled as the firm being located at multiple points on the line such that more predictions are in close vicinity. Alternatively, the same can be modelled as the firm still located at one point $e_i$ but the transportation cost shrinking—as a function of XAI level $\xi$—to $t(1 - \xi)$. At full XAI $\xi = 1$, the decision tree (Figure 2) explains all splits and the one-dimensional function (Figure 3) will have a large number of explained points ($\star$), eventually explaining all customer predictions.

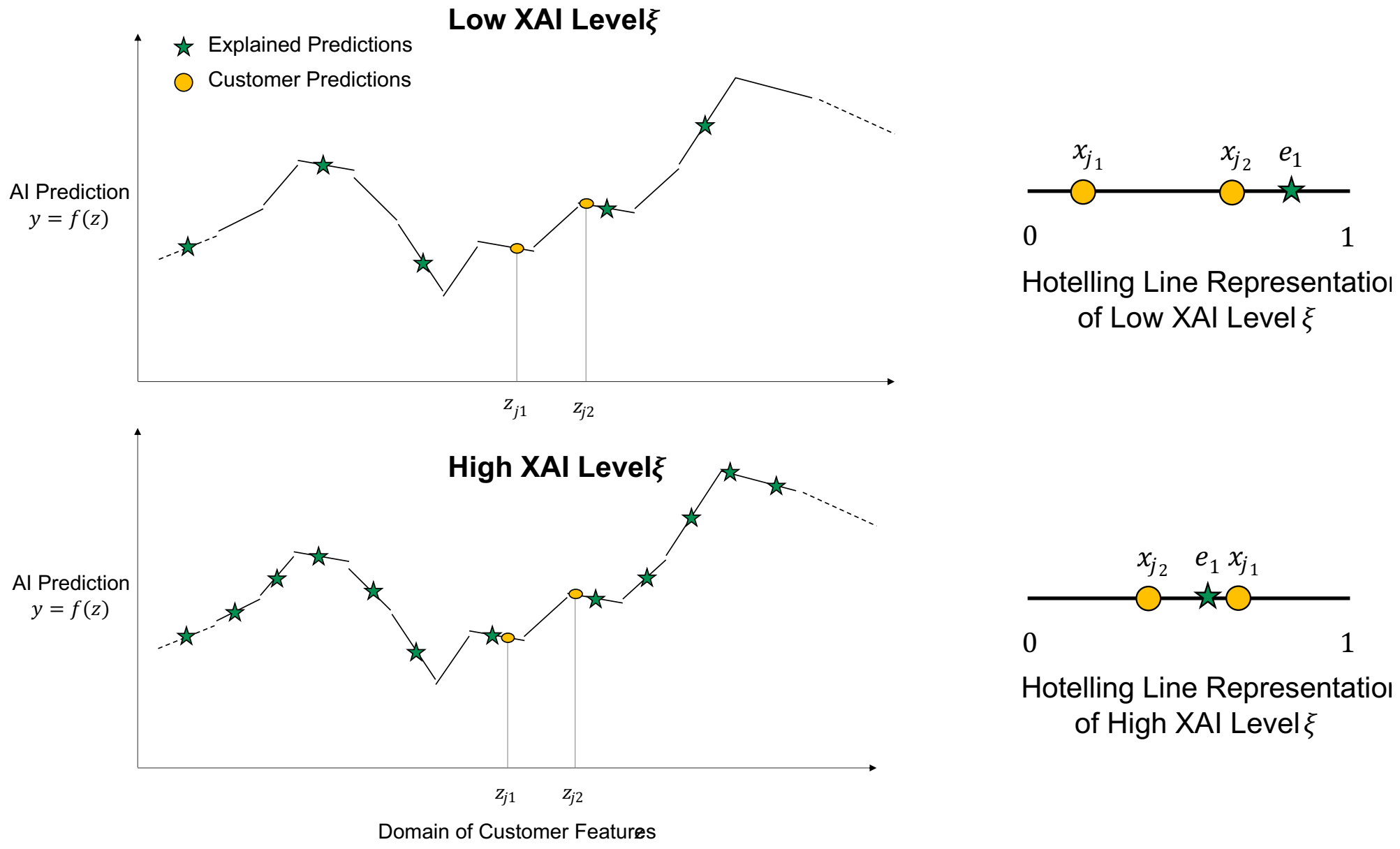

Figure 3. (Left) Firm $i$'s AI prediction model $y = f(z)$ where $z$ is the input vector. $j_1$ and $j_2$ are two representative customers. Each customer gets a prediction $y_j$ and an explanation. At low XAI level, the explanation is local for customer $j_2$ but not for $j_1$. At high XAI level, the explanation is local for both $j_1$ and $j_2$. (Right) Equivalent Hotelling line representation for low and high XAI levels. All explained points collapse into a single firm position $e_1$ on the Hotelling line while the customer position $x_j$ is relative to the closest explained point.

**XAI method $e$**: If firms use the exact same XAI method, they are co-located with each other, i.e., $e_1 = e_2$. When both firms offer the same XAI level $\xi$ (e.g., explaining 10 representative points on the one-dimensional function) but use different XAI methods $e_1 \neq e_2$ (e.g., explain a different set of 10 points), explanations offered by each firm better suit a different set of customer predictions. We analytically model two extremes: Same method ($e_1 = e_2 = 1/2$, denoted by 'min') and different methods ($e_1 = 0$ , $e_2 = 1$, denoted by 'max'[a]). Note that the intuitive examples in Figure 2 and Figure 3 are not exhaustive. XAI methods can differ in the set of explained predictions, the set of explained features, the way explanations are presented, or whether explanations are counterfactual (e.g., your chance of receiving the loan would be more than 50% if your salary were to increase by $5,000) vs. inferential (e.g., your loan was rejected because your salary was $4,000 less than average). In practice, consumers can be heterogeneous in their preference for one or the other XAI method based on any of these differences. We consciously do not model specific XAI methods to ensure that our model covers all current and future XAI methods.

In aggregate, customer $j$'s indirect utility from firm $i$'s quality, price, and explanation is as follows:

[a] An example of different XAI methods is when one firm uses SHAP and the other uses LIME.

$$u_{ij} = V_j + u_{ij}^q + u_{ij}^p + u_{ij}^e = V_j + \theta_j q_i - p_i - t(1-\xi_i)|x_j - e_i| \quad 3.1$$

where $V_j$ is the customer's reservation price or income, i.e., the intrinsic value that he has for the products in this market. We assume a covered market, so $V_j$ is large enough for all customers that $u_{ij}$ is always positive in equilibrium and every customer makes exactly one purchase[a]. The customer's utility from explanations $u_{ij}^e$ depends on the XAI level of firms $\xi_i$, the XAI methods $e_i$, and customer's preference for explanation $x_j$.

It is worth mentioning that product attributes (quality and explanations) must be orthogonal in our model, that is, $q_i$ and $\xi_i$ do not depend on each other. In some applications, such as autonomous vehicles and medical image diagnostics, quality may be contingent on the AI model's accuracy. On the other hand, the conventional wisdom used to assume a tradeoff between AI accuracy and interpretability (Adadi and Berrada, 2018), which violates the orthogonality of $q_i$ and $\xi_i$. That said, more recent research points to the contrary. (Rudin, 2019) argues that such a tradeoff is a "blind belief" and a "myth" that has no real data to support it. Rudin shows that in problems with structured data and meaningful features, there is often no significant difference in accuracy between more complex classifiers (deep neural networks, boosted decision trees, random forests) and much simpler[b] classifiers (logistic regression, decision lists) after pre-processing. Therefore, our model consciously does not incorporate any tradeoff between XAI and AI accuracy, and $q_i$ and $\xi_i$ remain orthogonal.

Moreover, some research looks at firm's tradeoff between XAI and the privacy of its AI model and trade secrets (Adadi and Berrada, 2018). It is shown that it may be possible to "steal" underlying AI models if AI input and outputs are provided (Barredo Arrieta et al., 2020; Orekondy et al., 2019). AI algorithms can also be subject to adversarial attacks that aim to confuse the model to lead it to a desired output by an adversary (Goodfellow et al., 2015). One could extrapolate from this and argue that firms may prefer not to use XAI because it may foster such issues. We are not aware of any research showing that XAI in practice has resulted in loss of a firms' intangible assets, led to copycats, or adversarial attacks. Thus, we exclude these factors from our model when considering firms' choice of XAI.

## 3.1. Market Structures

According to above, $(e_1, q_1)$ and $(e_2, q_2)$ represent the positions of firm 1 and firm 2 on the $q$–$x$ plane, respectively. At the same time, customers' preferences for quality and explanation are uniformly distributed over a unit square on the $\theta$–$x$ plane such that customer $j$ is represented by $(x_j, \theta_j) \in [0,1]^2$. Firm $i$'s demand (market share) $d_i$ is the area of this unit square that firm $i$ captures. Obviously, $0 \leq d_i \leq 1$ and $d_1 + d_2 = 1$.

---

[a] This, of course, makes use of the common assumption in the discrete-choice framework literature that income effects from price changes are negligible (Cunha et al., 2020), that is, income and prices are additive separable. Therefore, $V_j$ can be omitted from Eq. 3.1 since it does not vary across products.

[b] And hence, more interpretable.

For the customer at $(x_j, \theta_j)$, the utility from the products offered by firm 1 and firm 2 is $u_{1j}$ and $u_{2j}$, respectively, where[a]:

$$u_{1j} = \theta_j q_1 - t(1-\xi)|x_j - e_1| - p_1;$$
$$u_{2j} = \theta_j q_2 - t(1-\xi)|x_j - e_2| - p_2. \qquad 3.2$$

Firm 1's demand $d_1$ are customers for whom $u_{1j} > u_{2j}$. Similarly, firm 2's demand $d_2$ are those for whom $u_{2j} > u_{1j}$. The set of customers who are indifferent between making a purchase from firms 1 and 2 is called the *marginal customers*, found by setting $u_{1j} = u_{2j}$ and represented by the *indifference line* $\theta = \theta(x)$ that divides the unit square on the $x$–$\theta$ plane. The slope of $\theta(x)$ determines whether customers in general care more about explanations or quality[b], and leads to two math expressions for the equilibrium. Therefore, we follow (Neven and Thisse, 1989; Vandenbosch and Weinberg, 1995; Wattal et al., 2009) to separately analyze the possible markets formed by the indifference line.

**Lemma 1. Existence of Two Market Structures**

Depending on the slope and the intercept of the indifference line, only two market structures can have Nash equilibrium: the "explanation-dominated" market and the "quality-dominated" market, denoted by E and Q, respectively[c] (See Figure 4).

In an explanation-dominated market (or **market E** for short), what separates customers is their taste in explanations rather than quality (Figure 4, left). Since $q_1 \geq q_2$, customers on the left side of the indifference line are firm 1's demand ($d_1$) and customers on the right are firm 2's ($d_2$). Here, for any $\theta_j \in [0,1]$ there exists a customer $(x_j, \theta_j)$ who is indifferent about the firms' products. By contrast, in a quality-dominated market (or **market Q**), customers are separated based on their taste in quality (Figure 4, right). Since $q_1 \geq q_2$, customers above the indifference line are firm 1's demand ($d_1$) and below it firm 2's ($d_2$). In this market, for any $x_j \in [0,1]$ there exists a customer $(x_j, \theta_j)$ who is indifferent about the firms' products.

[a] $V_j$ is omitted due to the previous footnote.

[b] The diagonal line $\theta(x) = x$ represents complete indifference between quality and explanation.

[c] Proof in Appendix A1.

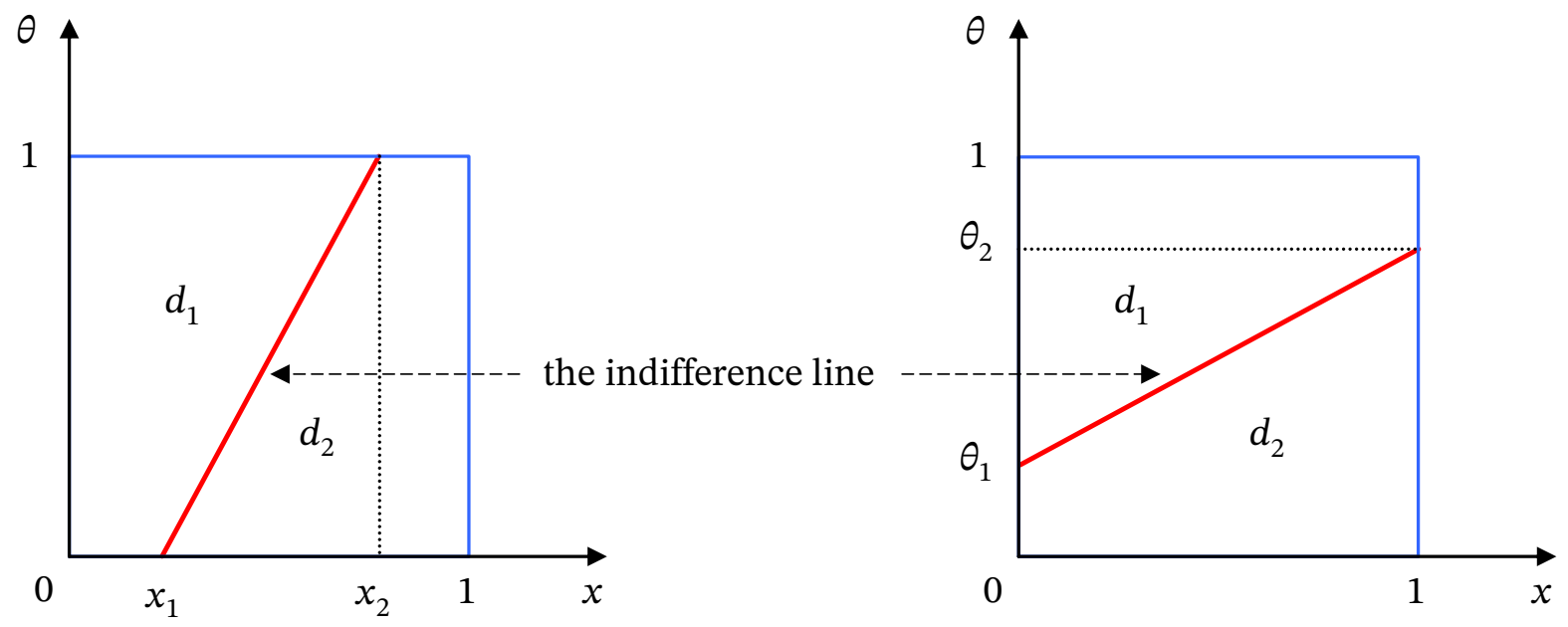

Figure 4. (Left) Explanation-dominated market. (Right) Quality-dominated market. The $x$- and $\theta$- axes represent customer preferences for explanations and quality, respectively.

## 3.2. The Policy-Maker's Problem

The policy-maker's goal is to set an XAI level $\xi$ for both firms such that the *total welfare* $W_t$ is maximized. $W_t$ is composed of firm profits and consumer utility, which is calculated by integrating Eq. 3.1 over $d_1$ and $d_2$.

We consider two policy setups: (1) **Optional** XAI, (2) **Mandatory** XAI. In mandatory XAI, the policy-maker sets the XAI level $\xi$ and enforces this policy. To do that, the policy-maker calculates firm profits and consumer utility— knowing that both firms offer XAI—and finds the optimal $\xi$. In optional XAI, the policy-maker sets the XAI level $\xi$ as a guideline, but firms are free to choose whether they will offer XAI at this level or they will offer no XAI at all. The policy-maker must predict firms' equilibrium strategies and set optimal $\xi$ accordingly. Through the choice of $\xi$, the policy-maker can shape the market structure, and hence firms' equilibrium strategies. We will also discuss scenarios where for some values of $\xi$, both market structures E and Q are possible at the same time. Firms will choose the market that yields higher profits, but in case they have conflicting interests, there will be no equilibrium. The policy-maker must avoid this situation by choosing $\xi$ properly.

## 3.3. The Firms' Problems

Firm $i$'s problem is to maximize its profit function $\pi_i$ which is defined as:

$$\pi_i = p_i d_i - \beta q_i^2. \qquad 3.3$$

Firms face two situations:

- Regulated XAI level: The policy-maker sets $\xi$ for both firms.
- Unregulated XAI level: Firms freely choose their own $\xi_i$.

Under regulated XAI level, the policy-maker sets $\xi_1 = \xi_2 = \xi$, so $\xi$ is endogenous for the policy-maker and exogenous for firms. In this situation, each firm $i$ has to maximize Eq. 3.3 by choosing the optimal $q_i^*$ and $p_i^*$. The third decision that firms must make has to do with offering XAI. As mentioned before, under optional XAI, each firm must choose whether or not it offers XAI at level $\xi$ set by the policy-maker. But under mandatory XAI, both firms are forced to offer XAI at level $\xi$.

Under unregulated XAI level, there is no policy-maker involved and each firm $i$ maximizes Eq. 3.3 by finding the optimal $\xi_i^*$, $q_i^*$, and $p_i^*$. This setting is not a primary focus of our paper and is only briefly discussed in Section 6 because we foresee the policy-maker taking an active role, even if for political or PR reasons.

## 3.4. The Games

The choice of XAI method ($e_i$) is a long-term strategy which often needs multiple levels of approval within the organization and with the regulators. The way XAI is presented also shapes how customers interact with the firm's product. Thus, changing $e_i$ requires significant investment in product repositioning. Thus, we treat firms' locations on the Hotelling line as exogenous decisions made before the start of the game. Moreover, firms incur zero technical cost of implementing XAI. The reason is that cutting-edge, model-agnostic XAI algorithms (e.g., SHAP, LIME) are now readily available as free and open-source packages.

We discuss the necessary setup for game solution for the optional XAI regulation where the policy-maker sets XAI level $\xi$. The game solutions for mandatory XAI will be simplified version of the same. There are two market structures E and Q under two firm differentiations 'max' and 'min'[a]. So, there are 4 games to analyze. Table 1 illustrates 3 equilibrium scenarios mentioned in Section 3.2 for each of these 4 games.

Table 1. Labels for 4 games and 3 equilibrium scenarios that could happen in each game.

| | | Firms' Horizontal Differentiation | | | |
|---|---|---|---|---|---|
| | | max | | min | |
| | | Market Structure | | Market Structure | |
| | | E | Q | E | Q |
| How ManyFirms Offer XAI? | 0 | maxE0 | maxQ0 | minE0 | minQ0 |
| | 1 | maxE1 | maxQ1 | minE1 | minQ1 |
| | 2 | maxE2 | maxQ2 | minE2 | minQ2 |
| | | Game 1 | Game 2 | Game 3 | Game 4 |

Each game has three sequential stages:

- **Stage 0:** Firms simultaneously decide whether they will offer XAI $\xi_i \in \{\xi, 0\}$;
- **Stage 1:** Firms simultaneously choose their quality levels $q_i$;
- **Stage 2:** Firms simultaneously choose their prices $p_i$.

[a] As a reminder: 'max': $e_1 = 0, e_2 = 1$, 'min': $e_1 = e_2 = 1/2$.

The intuition behind the three-stage game structure is the fact that prices are more flexible than quality in the short-term (Cunha et al., 2020). Likewise, product quality (e.g., through the AI model's accuracy) is more flexible than XAI strategy. Thus, decisions in early stages can be viewed as the firm's long-term strategy while subsequent stages involve progressively shorter-term decisions.

Stage-0 equilibrium decisions could be one of the following:

- No firm offers XAI: $(\xi_1^*, \xi_2^*) = (0{,}0, )$
- One firm offers XAI: $(\xi_1^*, \xi_2^*) = (\xi, 0)$ or $(0, \xi)$
- Both firms offer XAI: $(\xi_1^*, \xi_2^*) = (\xi, \xi)$

For example, with $(\xi_1^*, \xi_2^*) = (\xi, 0)$ firm 1 offers explanations at equilibrium while firm 2 does not. Firms' payoff functions $\pi_1, \pi_2$ and strategies are common knowledge. We use backward induction to solve for pure strategy Nash equilibria.

# 4. Regulated XAI (Exogenous $\xi$)

## 4.1. Optional XAI

First, we present a summary of stage-0 equilibrium decisions in Table 2. Then we discuss the equilibria in markets Q and E. Having found the equilibria, we will then address the problem of policy-making for optional XAI in Section 4.1.3.

Table 2. Stage-0 equilibrium decisions $(\xi_1^*, \xi_2^*)$[a]

| | | Market Structure | |
|---|---|---|---|
| | | Explanation-Dominated (E) | Quality-Dominated (Q) |
| Horizontal Differentiation | max | $\begin{cases} (\xi, 0) & \xi > \xi_* \\ (0,\ 0) & \text{else} \end{cases}$ | $(\xi, \xi)$ |
| | min | $(\xi, 0)$ | $(\xi, \xi)$ |

### 4.1.1. Equilibria in Quality-Dominated Markets (Q)

**Proposition 1.**

Under optional XAI, only the quality-dominated market has both firms offering explanations at level $\xi$ set by the policy-maker, and this is irrespective of firms' horizontal differentiation (See Table 2).

In market Q, we find that $(\xi_1^*, \xi_2^*) = (\xi, \xi)$ *weakly* dominates $(\xi_1^*, \xi_2^*) = (0, 0)$ (See Table 3). Intuitively, this is due to symmetric $\xi$ and the fact that firms capture the entire Hotelling line in market Q[a]. Therefore, the misfit cost $t$ has no effect on the equilibrium[b].

[a] For the condition of each equilibrium, see Appendix A6.

Table 3. 4 identical scenarios in quality-dominated markets

| maxQ0 | maxQ2 | minQ0 | minQ2 |
|---|---|---|---|
| $p_1^* = \frac{4}{27\beta}, \quad p_2^* = \frac{2}{27\beta}, \quad q_1^* = \frac{2}{9\beta}, \quad q_2^* = 0, \quad \pi_1 = \frac{4}{81\beta}, \quad \pi_2 = \frac{2}{81\beta}$ | | | |

## 4.1.2. Explanation-Dominated Market (E)

### 4.1.2.1. Firms Using Different XAI Methods ('max')

Here we find that no firm has the incentive to offer explanations at low levels ($\xi < \xi_*$ where $\xi_* = 1 - 1/36\beta t$). That said, if the policy-maker sets $\xi > \xi_*$, then firm 1 (the high-quality firm) provides explanations while also offering a *higher* level of quality compared to a baseline in which no firm offers explanations. Firm 2 (the low-quality firm) responds by lowering its quality. Therefore, on one hand we have a quality gap which also grows in $\xi$ and contributes to greater profits. And on the other hand, explanations reduce the horizontal differentiation between the two firms on the Hotelling line, which should lower their profits. The net effect of these two forces is as follows:

**Lemma 2.**

Under different XAI methods, as long as firm 1 offers XAI at high levels ($\xi > \xi_*$), its profit increases with $\xi$ while firm 2's profit declines.

The intuition is that as $\xi$ increases, firm 1 (firm 2) increases (decreases) its quality level such that firm 1 always charges a higher price than firm 2. Therefore, firm 1 captures more customers while also increasing its price, whereas firm 2 loses its market share while lowering its price[c]. We show that with firm 1 offering XAI, firm 2 cannot do any better by adopting XAI, ruling out the $(\xi_1^*, \xi_2^*) = (\xi, \xi)$ strategy.

### 4.1.2.2. Firms Using Similar XAI Methods ('min')

The situation is different in this case. As seen in Table 2, firm 1 always finds it more profitable to offer explanations while firm 2 does not. Interestingly, firm 1 (the high-

[a] See Figure 13 in Appendix A2.

[b] One might consider a *fixed cost* $1\{\xi > 0\} \cdot c_{\text{XAI}}$ for XAI, say, due to implementing XAI for the first time. However, this would not change equilibrium results because Table 3 tells us that equilibrium price, quality, and profit are independent of $\xi$. In other words, $(\xi_1^*, \xi_2^*) = (\xi, \xi)$ still weakly dominates $(\xi_1^*, \xi_2^*) = (0, 0)$ in market Q.

[c] See Figure 14 (middle) in Appendix A2.

quality firm) *lowers* its quality as $\xi$ increases while firm 2 (the low-quality firm) does the opposite. Moreover, in contrast with Lemma 2, we have:

**Lemma 3.**

Under similar XAI methods, the profits of both firms increase as more explanations are asked by the policy-maker ($\partial\pi_i/\partial\xi > 0, i = 1, 2$).

The intuition is as follows: On one hand, profits increase with quality for both firms ($\partial\pi_i/\partial q_i > 0, i = 1, 2$), leading them to choose almost identical qualities if the cost of quality approaches zero[a], hence a reduction in the quality gap and profits. On the other hand, offering XAI introduces a measure of horizontal differentiation even though firms are located at the same point on the Hotelling line. This is because the increase in customer utility due to explanations is higher for customers who are farther away from the center than for those closer[b]. This is shown below:

$$\begin{aligned} &u_1^e = -tx(1-\xi); \\ &u_2^e = -tx; \\ &\quad \Rightarrow \Delta u^e \stackrel{\text{def}}{=} u_1^e - u_2^e = t\xi x. \end{aligned} \tag{4.1}$$

As $x$ gets larger, so does $\Delta u^e$, which enables the firm that offers explanations (firm 1) to capture customers located away from the center, while the other firm (firm 2) captures customers located close to the center. Thus, although firms are undifferentiated, XAI by one firm helps both firms earn higher profits.

Sections 4.1.2.1 and 4.1.2.2 can be summarized like so:

**Proposition 2.**

When firms use different (similar) XAI methods, XAI and quality are *complements* (*substitutes*).

### 4.1.3. Policy-Making for Optional XAI

The policy-maker aims to maximize the total welfare by choosing the optimal $\xi = \xi_{\text{policy}}^{\text{opt.}}$ for each value of $\beta t$, where $\xi_{\text{policy}}^{\text{opt.}}$ is the explanation level under optional XAI. Since firms' locations on the Hotelling line is exogenous, the policy-maker must devise an $\xi$ for 'max' and 'min' separately. The first step is to calculate $W_t$. If only one market (E or Q) is possible, then the policy-maker maximizes $W_t$ in that market w.r.t. $\xi$. But if two markets are possible at the same time, the strategic policy-maker calculates $W_t$ for each market and then chooses $\xi$ such that firms only choose his desired market and maximum total welfare is obtained.

Table 7 in Appendix A6 lists all equilibrium results and their conditions. According to the condition of each equilibrium, the $\beta t$–$\theta$ plane is divided into several regions as shown in Figure 5. While we have analytically derived $W_t$ for all equilibria[a], the cases of maxE1, minQ2 and minE1 are overly complex, making it analytically intractable to compare $W_t$ in different regions and calculate the optimal $\xi$. To tackle this problem, we run Monte-Carlo simulations on 40,000 points in the unit square in Figure 5. In doing so, we also find the markets that firms choose to form and areas where firms have conflicting incentives (shown in gray ⊗). In these regions, one firm prefers market E and the other prefers market Q, leading to no Nash equilibrium in the market.

---

[a] Which is not possible because of the equilibrium condition ($\beta t\xi > 2/9$).

[b] See Figure 14 (right) in Appendix A2.

In Figure 5, maximum total welfare is achieved at the points on the (thick) blue lines. Contrary to expectation, these graphs reveal the following unexpected result:

**Proposition 3.**

Under optional XAI, irrespective of firms' differentiation, policy-makers who aim to maximize total welfare should not always require full explanations.

This finding is because for some values of $\beta t$, high XAI level leaves the firms indecisive about the market they seek to form. This can be verified analytically for the 'min' case:

**Lemma 4.**

Under optional XAI, when firms use the same XAI method and both markets E and Q are possible (Figure 5, right, gray area ⊗), the high-quality firm always earns more in market E than it does in market Q, but the opposite is true for the low-quality firm. Choosing $\xi^{\text{opt.}}_{\text{policy}}$ in this region, then, leads to no Nash equilibrium[b].

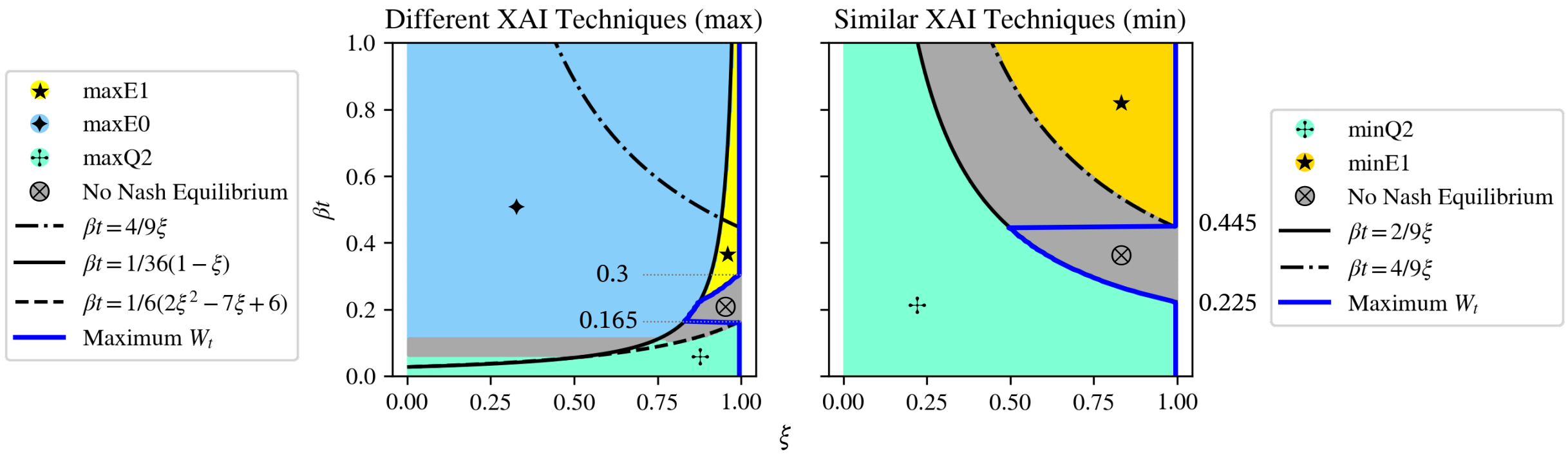

Figure 5. Different regions on the $\beta t$–$\xi$ plane and the equilibria that they support under optional XAI

## 4.2. Mandatory XAI

If it is mandatory for firms to offer XAI at level $\xi$ set by the policy-maker, Table 1 reduces to only 4 scenarios shown in Table 4.

[a] See Appendix A9.

[b] Proof in Appendix A8.

Table 4. The scenarios in which both firms offer XAI

| | Scenario | | | |
|---|---|---|---|---|
| | maxE2 | maxQ2 | minE2 | minQ2 |
| Is it equilibrium in optional XAI? | No | Yes (in maxQ) | No | Yes (in minQ) |

As we saw in Section 4.1.1, maxQ2 and minQ2 in Table 4 are equilibrium in quality-dominated markets. This implies that even without policy enforcement, both firms offer XAI in these scenarios, regardless of the $\xi$ level. Moreover, scenario minE2 is discarded because the indifference line is horizontal and firms engage in a price war while racing to the bottom of quality, rendering this scenario impossible to sustain. So, there is only scenario maxE2 that is possible but is not equilibrium by default.

The condition for each scenario, total welfare $W_t$, and its derivative w.r.t. $\xi$ are presented in Table 5. Unlike the case of optional XAI, here $W_t$ expressions are analytically tractable and there is no need for Monte-Carlo simulations.

Table 5. The 3 possible scenarios under mandatory XAI

| | maxE2 | maxQ2 | minQ2 |
|---|---|---|---|
| $W_t$ | $\dfrac{t(\xi-1)}{4}+\dfrac{1}{36\beta}$ | $\dfrac{t(\xi-1)}{2}+\dfrac{1}{27\beta}$ | $\dfrac{t(\xi-1)}{4}+\dfrac{1}{27\beta}$ |
| $\partial W_t/\partial\xi$ | $t/4>0$ | $t/2>0$ | $t/4>0$ |
| Condition | $\beta t>\dfrac{1}{6(2\xi^2-7\xi+6)}$ | $\beta t<\dfrac{4}{9\xi}$ | $\beta t<\dfrac{4}{9\xi}$ |

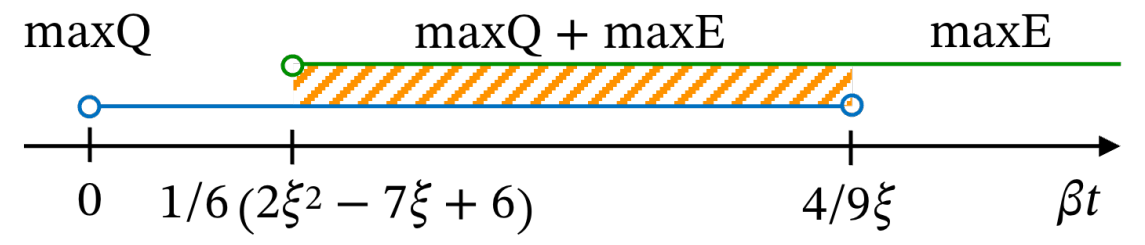

In Table 5, the greatest total welfare among all scenarios belongs to maxQ2, followed by minQ2 and then maxE2, and this holds for all $\xi\in[0,1]$.

**Lemma 5.**

When firms are forced to offer XAI and they use the same XAI method ('min'), only scenario minQ2 is possible. As such, the policy-maker sets $\xi_{\text{policy}}^{\text{mand.}}=\min\{4/9\beta t\,,1\}$ to maximize $W_t$ in this scenario, where $\xi_{\text{policy}}^{\text{mand.}}$ is the XAI level chosen by the policy-maker under mandatory XAI to maximize total welfare.

On the other hand, two scenarios are possible when firms use different XAI methods ('max'). Looking at the conditions of maxE2 and maxQ2, we find that both scenarios are possible for a range of $\beta t$. In this interval, market Q yields a higher total welfare than market E and is thus preferable by the policy-maker[a]. But firms select the market that maximizes their profit, which may not necessarily maximize $W_t$. Therefore, the policy-maker must choose $\xi_{\text{policy}}^{\text{mand.}}$ such that only one market is possible and $W_t$ is maximized. Based on the conditions of each market, the $\beta t$–$\xi$ plane is split into several regions as illustrated in Figure 6[b] and we obtain the following surprising finding:

**Proposition 4.**

When firms are forced to offer XAI, irrespective of firms' horizontal differentiation ('max' or 'min'), mandating firms to offer full-explanations may actually make everyone worse off, including the very customers that the policy aims to support.

In the case of 'min', it is because there will be no equilibrium if $\xi_{\text{policy}}^{\text{mand.}} > 4/9\beta t$. And in the case of 'max', it is because with $\xi_{\text{policy}}^{\text{mand.}} = 1$, firms may end up in the red region (!) in Figure 6 (left) in which $\pi_1, \pi_2 < 0$ and firms leave the market, resulting in $W_t = 0$.

A question worth asking at this point is: Why is $\xi_{\text{policy}}^{\text{mand.}} < 1$ for large $\beta t$? The answers involves the red region (!) in Figure 6 and is as follows[c]: We know that $(\xi_1^*, \xi_2^*) = (\xi, \xi)$ is not the equilibrium strategy of maxE by default[d]. Mandating XAI, then, makes both firms worse off compared to their equilibrium strategy. The intuition is that under the mandatory regulation, as $\xi$ increases, the horizontal differentiation between the firms on the Hotelling line decreases because each firm's horizontal differentiation to the other is $t(1-\xi)|e_1 - e_2|$. This in turn leads to price competition and lower profits. Worse yet, there is no quality gap between the firms to increase the profits ($\Delta q = 0$). The result is a decline in the profits as shown in Figure 7.

In Figure 6 (right), $\xi_{\text{policy}}^{\text{mand.}}$ converges to 1 as $\beta t \to \infty$, suggesting that when the cost of quality and misfit cost are high enough, the policy-maker can require more explanations. Overall, our results in this Section and before point to the significance of $\beta$ (hence the marginal cost of quality) and $t$ (hence the misfit cost) in devising XAI policies, both in optional and mandatory XAI.

[a] See the expressions of $W_t$ in Table 5.

[b] See Appendix A10 for more details and mathematical expressions of regions and curves.

[c] See also Appendix A10.

[d] In Section 4.1.2, we found the equilibrium strategies to be $(\xi_1^*, \xi_2^*) = (0, 0)$ for $\xi < \xi_*$ and $(\xi_1^*, \xi_2^*) = (\xi, 0)$ for $\xi > \xi_*$.

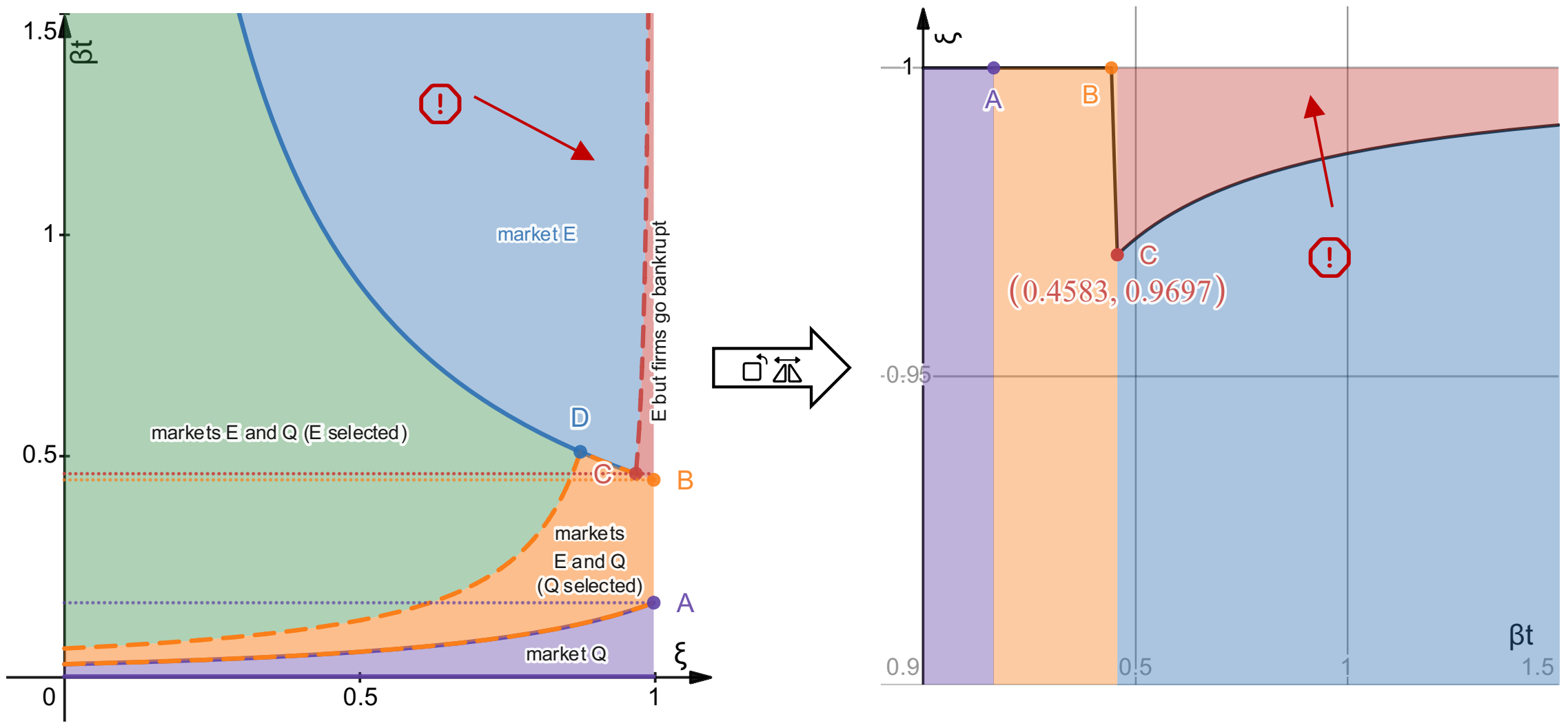

Figure 6. (Left) Different regions on the $\beta t$–$\xi$ plane and the markets that they support under mandatory XAI and 'max'. (Right) The policy $\xi_{\text{policy}}^{\text{mand.}}$ that guarantees maximum total welfare

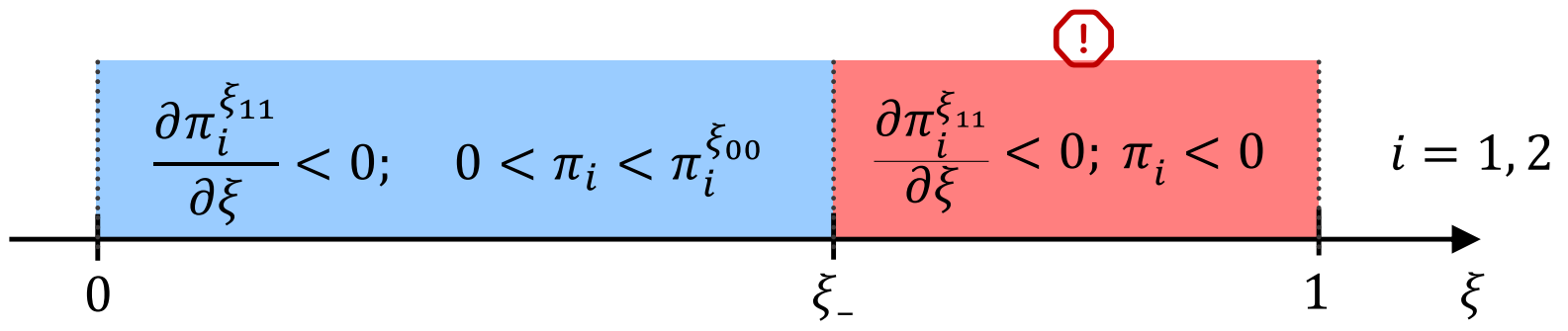

Figure 7. Under mandatory XAI, firms are worse off if they use different XAI methods in market E

# 5. Regulatory Insights

## 5.1. Mandatory vs. Optional XAI

In light of the results of Section 4, a natural question is: Which one—optional or mandatory XAI—leads to higher total welfare? The answer may not be obvious at first glance. The total welfare does not simply depend on the explanation level; it also depends on the number of firms who offer XAI as well as price and quality levels. Take the case of 'max' under high $\beta t$ for instance. On one hand, under optional XAI, the policy-maker asks for full explanations ($\xi_{\text{policy}}^{\text{opt.}} = 1$) but *only one* firm will choose to offer XAI (see Figure 5). On the other hand, if XAI is mandatory, the policy-maker asks for partial explanations ($\xi_{\text{policy}}^{\text{mand.}} < 1$) but *both* firms must comply (see Figure 6). Thus, one setting has higher explanation level while the other has a greater number of firms that offer XAI. The net effect is illustrated in Figure 8.

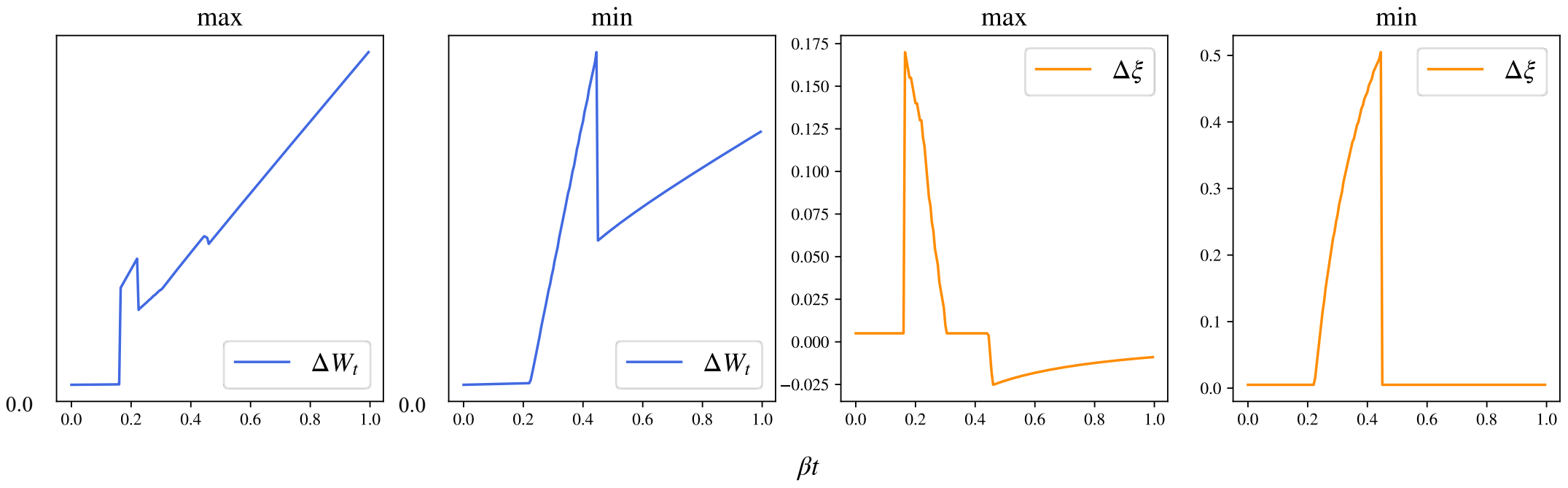

Figure 8. The gaps between total welfare and explanation levels of mandatory and optional XAI. $\Delta W_t := W_t^{\text{mand.}} - W_t^{\text{opt.}}$, and $\Delta\xi := \xi_{\text{policy}}^{\text{mand.}} - \xi_{\text{policy}}^{\text{opt.}}$.

First, our findings support the intuition that in general, mandating XAI makes the society as a whole better off ($\Delta W_t \geq 0$). That said, we have:

**Proposition 5.**

Comparing optional and mandatory XAI, there is no additional total welfare from mandating XAI ($\Delta W_t = 0$) when $\beta t$ is small.

The intuition is that small cost of misfit $t$ suggests that the market does not care so much about explanations and demands are separated based on customers' taste in quality rather than explanations (quality-dominated market). As such, explanation levels have no effect on firms' profits, even when $\xi$ is 1. Contrary to popular belief, we also find that:

**Proposition 6.**

Under mandatory and optional XAI, full-explanations are not always necessary to achieve maximum total welfare.

In fact, mandatory policies may actually require *less* explanation levels than optional policies (Figure 8, third from left), because while optional XAI level can be 1 for $\beta t > 4/9$, mandatory XAI level must remain less than 1 to allow for market equilibrium[a].

## 5.2. Other Policy-Making Objectives

In Section 3.2 we mentioned that the policy-maker's goal is to maximize the total welfare of the society. But in practice the policy-maker may find it challenging to measure the total welfare. For political or PR reasons, the policy-maker may seek to maximize an objective that is easier to measure and publicize, such as the total *number of firms* that opt-in to offer XAI, the *average level of explanations* received by customers, or XAI *fairness*. These are discussed next.

[a] As mentioned at the end of Section 4.2.

#### 5.2.1. Number of Firms that Offer XAI and Average XAI Level

According to Figure 9 (right), when firms have the same XAI method, the number of firms that offer XAI is *not* monotonically increasing in $\xi$. Similarly, the average XAI level $\xi_{\text{avg.}}$ received by customers[a] is *not* always monotonically increasing in $\xi$ (e.g., in Figure 9, right).

**Proposition 7.**

Asking for full explanations is not always the best regulatory decision even when the policy-maker's objective is the total number of firms that offer XAI or the average XAI level.

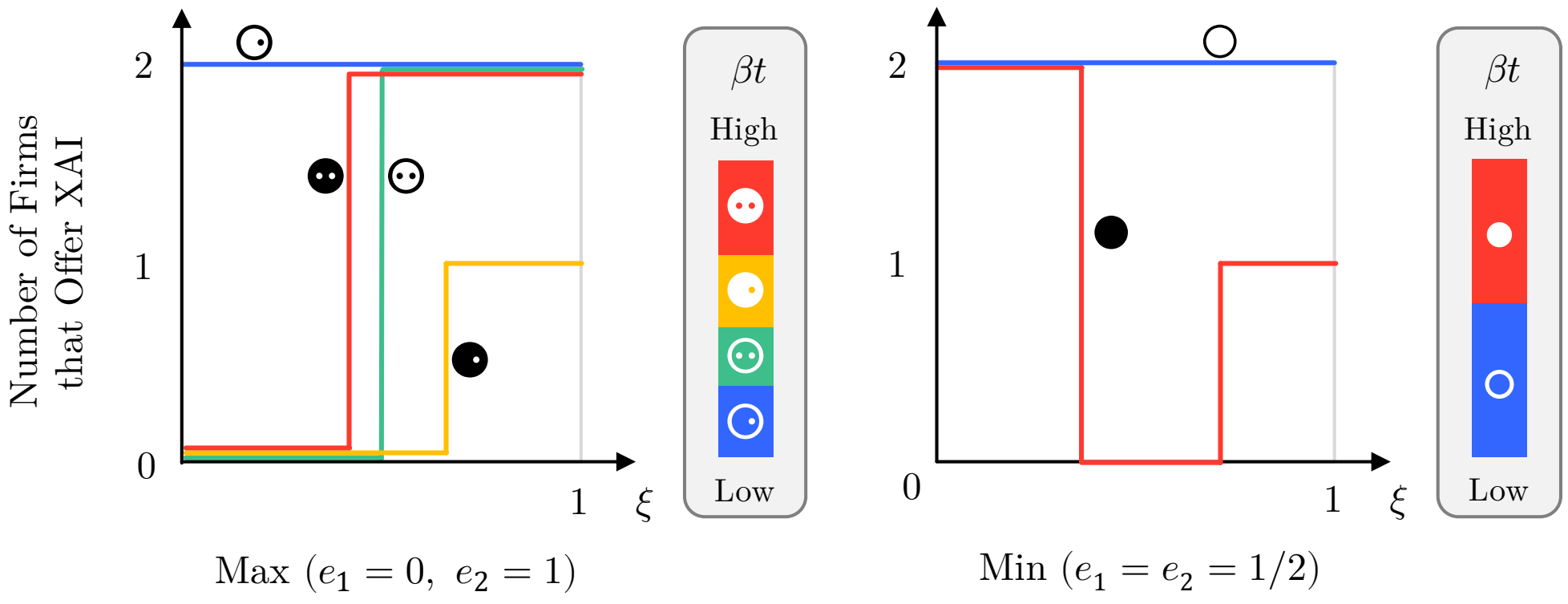

Figure 9. The number of firms that offer XAI as a function of $\xi$

#### 5.2.2. XAI Fairness

While XAI can help identify potential sources of AI unfairness (Zhou et al., 2022), perhaps less obvious is the potential unfairness in XAI itself through explanations *delivery*. To our knowledge, no previous studies have investigated this aspect of XAI. In order to fill this gap, here we introduce two dimensions of XAI fairness[b] and discuss the policy-maker's strategy if his objective is XAI fairness. Notice that XAI fairness is not the same as AI fairness: With model-agnostic XAI methods, XAI fairness could be violated even if the underlying AI models are fair.

##### 5.2.2.1. Vertical XAI Fairness

Take any customer $j$ located at $(x_j, \theta_j)$ in the market[c]. We define *vertical fairness* as the situation where all customers $(x_j, \theta_k), \theta_k \in [0, 1]$ receive the same level of explanations as each other[a]. This happens when both firms offer equal XAI levels[b] ($\xi_1 = \xi_2$). The intuition behind vertical fairness is as follows: Consider two customer groups—protected (historically marginalized) and unprotected (historically advantaged). In our bank loan example, if product quality is tied to the AI model's confusion matrix[c], then according to anecdotal evidence, the protected group may prefer lower False Positives[d] while the unprotected group prefers something else. This implies different preferences for qualities and leads to members of each group clustered together on the $\theta$-axis.

[a] $\xi_{\text{avg.}} \overset{\text{def}}{=} d_1\xi_1 + d_2\xi_2$

[b] See Figure 16 in Appendix A12.1 for the illustration of our notion of XAI fairness.

[c] As in Figure 4.

To attain vertical fairness, the policy-maker sets an explanation level $\xi_{\text{fair}}$ and forces both firms to offer XAI at this level, i.e., $\xi_1 = \xi_2 = \xi_{\text{fair}}$. This is identical to mandatory XAI when $\xi_{\text{fair}} = \xi^{\text{mand.}}_{\text{policy}}$. Our analysis in Section 4 shows that optional XAI, too, can provide vertical fairness under maxE0, maxQ2, and minQ2 scenarios[e]. However, in optional XAI, there exist conditions[f] under which there is a tradeoff between vertical fairness and total welfare $W_t$. Therefore, only mandatory XAI can maximize total welfare and achieve vertical fairness simultaneously (see Figure 10).

#### 5.2.2.2. Horizontal XAI Fairness

*Horizontal fairness* refers to the situation in which all customers on the Hotelling line receive the same XAI (dis)utility. This notion of XAI fairness is similar to the "equal outcome" definition of AI fairness (Fu et al., 2022; Hardt et al., 2016) and can only happen when firms offer full explanations ($\xi_1 = \xi_2 = 1$)[g]. Intuitively, horizontal fairness addresses the situation where the AI predictions for the protected and unprotected groups may be best explained by different features, e.g., employment or education for one group but credit history and purpose of loan for another group. Thus, members of the protected group will be closely clustered together on the Hotelling line (say, on the right extreme), distant from the unprotected group (say, in the middle).

One might think that setting $\xi_{\text{fair}} = 1$ attains vertical *and* horizontal fairness as well as maximum total welfare under mandatory XAI. However, for $\beta t > 4/9$, setting $\xi_{\text{fair}} = \xi^{\text{mand.}}_{\text{policy}} = 1$ is simply impossible[h], and this leads to a rather surprising finding:

---

[a] This does *not* imply that everyone on the $x$-axis receives the same XAI (dis)utility.

[b] See Appendix A12.1 for proof.

[c] For a binary classification task, the confusion matrix is a $2 \times 2$ table including the number of True Positives, True Negatives, False Positives (type I error), and False Negatives (type II error).

[d] Where "positive" in this example means "high chance of default", hence "reject".

[e] See Table 11 in Appendix A12.2. Also notice that even *unregulated* XAI may lead to symmetric XAI levels (See Proposition 9 and Proposition 10).

[f] See Appendix A12.3.1.

[g] We must have $u^e_{1j} = u^e_{2j}\ \forall x_j \in [0,1]$. Thus, $\xi_1 = \xi_2 = 1$ and $u^e_{1j} = u^e_{2j} = 0$.

[h] See Appendix A12.3.2.

**Proposition 8.**

One cannot guarantee horizontal fairness for all $\beta t$, even if the policy-maker were to forego maximum total welfare.

If even mandatory XAI cannot guarantee horizontal fairness, could optional XAI be as good as mandatory XAI? At first, it appears to be the case because unlike the case of vertical fairness, there are no tradeoffs[a] between horizontal fairness and total welfare $W_t$ under mandatory and optional XAI. That said, Figure 10 shows that mandatory XAI is still preferred in terms of XAI fairness that it provides. In other words:

**Theorem 1.**

If the regulator's objective is XAI fairness, then his optimal strategy is a policy that *mandates* XAI. This policy guarantees vertical fairness and attains horizontal fairness when $\beta t < 4/9$. The said policy also maximizes total welfare $W_t$.

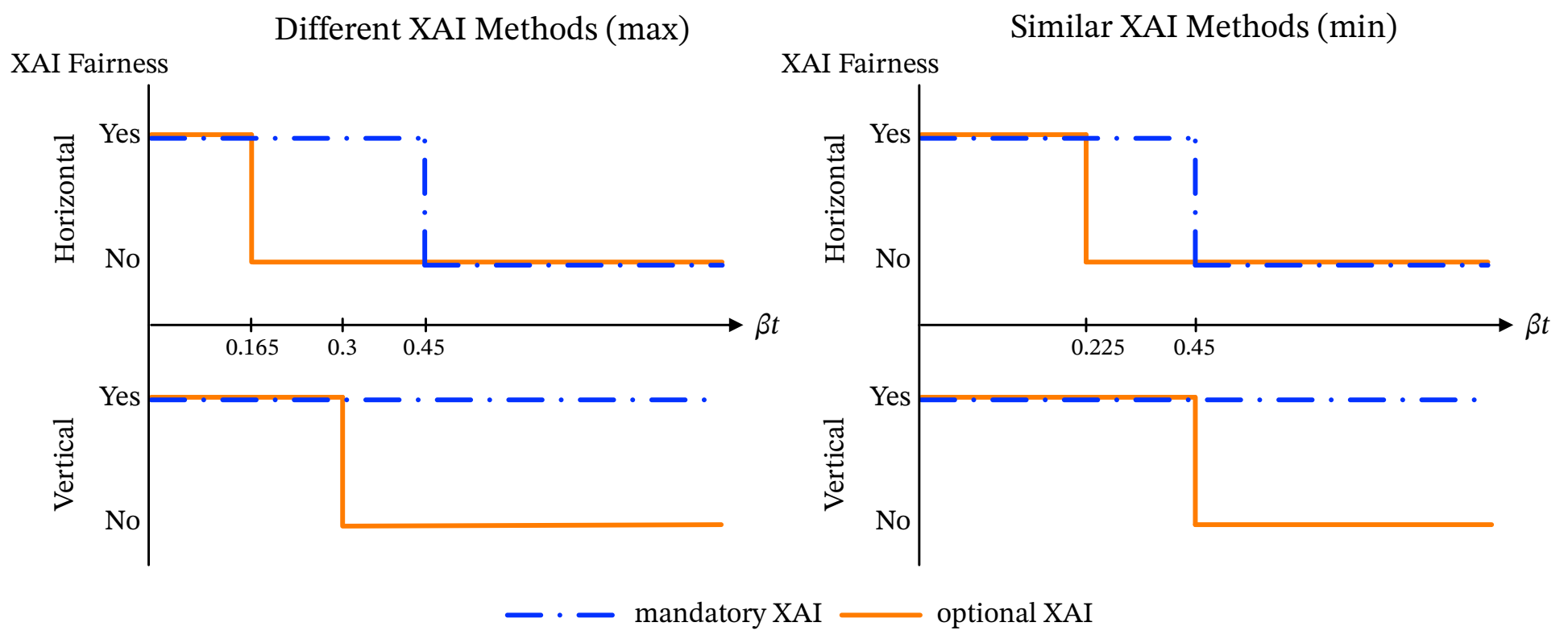

Figure 10. Horizontal and vertical XAI fairness under mandatory and optional XAI policies

# 6. Unregulated XAI (Endogenous $\xi$)

Until now, each firm's XAI decision was either a binary option of $\{0, \xi\}$ (in optional XAI) or fixed to the policy-maker's choice of XAI level (in mandatory XAI). A natural question is: What would firms do if left unregulated? In our model, this would correspond to firms' choices of XAI levels $(\xi_1, \xi_2)$. To answer this, we extend the method in (Wattal et al., 2009) by deriving equilibrium results when $\xi$ is endogenous for firms and not necessarily symmetric[b]. One interesting finding is as follows:

**Proposition 9.**

In an unregulated market, when firms adopt different XAI methods in market E, an *asymmetric* equilibrium is not sustainable.

[a] See Appendix A12.3.2.

[b] We have provided more detailed results in Appendix A13.

In particular, firms always mirror each other's XAI levels[a] such that $\xi_1 = \xi_2 = 1 - 1/36\beta t$. This implies that $\xi_1, \xi_2$ will never be 1[b]. Markedly, this result indicates that **firms choose to hide some information in their XAI outputs even though XAI does not cost anything**. As the misfit cost $t$ increases and the market cares more about explanations, $\xi_1, \xi_2$ approach 1. Unfortunately, analytical derivations for market Q are expectedly more complex than the regulated case presented in Section 4; we are only able to attain closed form solutions for the E-market.

To tackle this problem, we use a **simplified model** as follows: Firms have a binary choice of quality $\{0, q\}$ and price $\{0, p\}$ instead of the continuous choice in the full model. Moreover, firms continue to have the full choice of XAI levels, i.e., $\xi_1, \xi_2 \in [0, 1]$. In this simplified model, once again we find that market E is only viable when firms are differentiated in XAI method. Besides, firms' choice of XAI levels is always symmetric, $\xi_1^* = \xi_2^*$, and can be less than 1. This is all consistent with the full model. But now using the simple model we can examine the Q-market and once we have the full equilibrium characterization, we can compare the unregulated and regulated settings[c]. Table 6 summarizes the equilibrium XAI levels of our simplified model:

Table 6. Equilibrium XAI levels in the simplified model under regulated and unregulated XAI

| | | Regulated XAI | | Unregulated XAI |
|---|---|---|---|---|
| | | Optional | Mandatory | |
| Horizontal | max | $(\xi_1^*, \xi_2^*) = \left(\xi_{\text{policy}}^{\text{opt.}^1}, 0\right)$ | $\xi_{\text{policy}}^{\text{mand.}} = 1 - \eta/2$ | $(\xi_1^*, \xi_2^*) = 1 - \eta/2$ |
| | min | $(\xi_1^*, \xi_2^*) = \left(0, \xi_{\text{policy}}^{\text{opt.}^2}\right)$ | No feasible $\xi_{\text{policy}}^{\text{mand.}}$ | $(\xi_1^*, \xi_2^*) = (1 - \delta, 1)$ |

$$\xi_{\text{policy}}^{\text{opt.}^1} = \min\{1, 2 - \eta\},\ \xi_{\text{policy}}^{\text{opt.}^2} = \delta$$

where $\eta \stackrel{\text{def}}{=} 1/2\beta t \times p/q$ and $\delta \stackrel{\text{def}}{=} 4q/t \times (1 - \beta q^2/p)$. An interesting finding is:

**Theorem 2.**

The equilibrium XAI level may be less than full-explanation ($\xi < 1$) in unregulated and regulated (optional or mandatory XAI) markets.

When both mandatory and unregulated XAI are possible (the 'max' case), one might argue in favor of mandatory XAI as it ensures vertical XAI fairness by enforcing firms

---

[a] But while mirroring each other's XAI *levels* leads to a symmetric equilibrium in this market, mirroring each other's XAI methods results in no equilibrium at all. See Appendix A13.

[b] It turns out that $\xi = 1 - 1/36\beta t$ is the same as the line $\beta t = 1/36(1 - \xi)$ on Figure 5 (left).

[c] We will additionally require that the policy-maker (when imposing mandatory or optional XAI) wants to ensure that both firms enter the market. This last requirement helps because the binary choice of price may not be realistic in a monopoly.

to offer an identical XAI level, even if it is not full explanation. That said, Table 6 shows that:

**Proposition 10.**

Unregulated XAI, too, can lead to symmetric XAI, and thus vertical XAI fairness.

In fact, the unregulated setting in 'max' is exactly the same as mandatory XAI—symmetric in price, quality, and XAI level. This is consistent with the indicative results from the full model discussed above. But contrary to the 'max' case, we find that no symmetric equilibrium is sustainable when firms use the same XAI methods ('min').

Overall, the results of our simplified model indicate that compared to regulated XAI, unregulated XAI may in fact provide better total welfare, total consumer utility, and average XAI level[a]. While we cannot analytically make the unregulated and regulated comparison for the full model, the results with the simple model are unfavorable to a regulated setting because firms are unable to differentiate on price and quality to the same extent. Therefore, the results from the simple model are more relevant to markets where price and quality are relatively standardized, e.g., AI-based lending where the interest rates may be pegged across the market.

# 7. Conclusion

Our paper provides several insights for policy-makers and managers. First, policy-makers in the field of XAI should pay attention to differentiation of firms in terms of XAI methods, market structure (explanation- vs. quality-dominated), and cost of quality and misfit explanations. Firms may develop their own XAI algorithms in-house or customize existing packages to suit their needs. Whether firms differentiate or not in terms of the XAI method, warrants a different regulatory treatment. Another aspect of policy-making is the market structure that is formed as a result of the policy and firms' decisions. This study shows that in some cases, both markets are possible and firms have to choose one of them. Policy-makers should take this fact into account and carefully choose the required XAI level such that firms pick the market that yields higher total welfare. Sometimes under optional XAI policies, firms are indecisive about the market that they want to form and policy-makers should choose the required XAI level *low enough* that both firms choose the same market. In fact, one of our results is that high levels of explanations even under optional XAI may actually lead to no Nash equilibrium in the market.

The results confirm the intuition that mandatory XAI generally makes the society better off, and this holds regardless of firms' XAI differentiation. That being said, **there are situations in which there is no additional benefit from mandating XAI**, especially when firms use similar XAI methods. This implies that policy-makers may need to focus on standardizing the level of explanations (optional XAI) instead of enforcing firms to offer XAI.

[a] See Appendix A13.3.

We find that requiring full explanations may actually make firms and consumers worse off, and this result holds for both optional and mandatory XAI policies. In fact, this is more of an issue under mandatory XAI. In this situation, policy-makers may have to settle for partial explanations, which means that AI models will remain partly black box. By contrast, optional XAI in differentiated markets may permit full explanations, albeit only one firm offering XAI. Therefore, we show that **there exists a tradeoff between maximizing the total welfare of the society (through mandatory XAI) and requiring full explanations (through optional XAI)**. We believe that this tradeoff is particularly important in differentiated markets because firms often develop their own XAI methods which are different from the competition (Bhatt et al., 2019).

We investigate several other policy-making objectives, including number of firms that offer XAI, average XAI level received by consumers, and XAI fairness. The results in these cases show that **requiring full explanations is not necessary to achieve the objective**, reaffirming our previous findings when the objective was maximizing total welfare. In doing so, we also contribute to the literature on fairness by providing two novel definitions of XAI fairness—vertical and horizontal fairness, the latter being similar to the "equal outcome" definition of AI fairness in the literature. To our knowledge, this is the first attempt in this area. We show that if the objective is XAI fairness, then policy-makers will naturally implement policies that are similar to **mandatory XAI, which not only maximizes total welfare, but also guarantees vertical fairness and attains horizontal fairness when $\beta t < 4/9$**. Moreover, we prove that horizontal XAI fairness is impossible to attain for all $\beta t$, implying that **different market segments (protected and unprotected groups) will likely experience unequal benefits from explanations**. We believe these results have ramifications for XAI policy-makers and firms. In particular, the impossibility of equal XAI benefits refutes efforts to promote XAI as the holy grail of AI fairness.

In sum, the takeaway message can be obtained from Proposition 3, Proposition 4, Proposition 7, and Theorem 1:

**Theorem 3.**

Whether firms use similar or different XAI methods, under mandatory and optional XAI there will always be situations where asking for full explanations is not the optimal policy-making decision, and this holds when the objective is total welfare, total number of firms that offer XAI, average XAI level, or XAI fairness.

An interesting observation in our results is that equilibrium solutions turn out to be expressed with $\beta t$ occurring together. The interchangeability of $\beta$ and $t$ highlights the underlying tradeoff between quality and explanations and the fact that these two must be considered jointly in XAI decisions. To support this notion, notice that the low-quality firm in our model never benefits from offering XAI unilaterally. In fact, even in market E it is the *high-quality* firm that may benefit from XAI. Ironically, this finding points to the **significance of quality as the main product attribute for firms that want to offer XAI.** Thus, managers using ML algorithms in their products should focus their attention on quality and treat it as a necessary condition for offering profitable XAI.

Finally, in the absence of policy-makers, it may appear that if technical threats of stolen IP, copycats, or adversarial attacks are eliminated and XAI methods are readily available for any AI algorithm, firms should maximize explanations as a lever to compete and charge higher prices. But our analysis suggests that **differentiated firms should actually avoid offering full explanations but *mirror* each other's XAI methods**. Moreover, a simplified version of our model shows that compared to regulated XAI (mandatory and optional), **unregulated XAI may in fact offer better total welfare, total consumer utility, and average XAI levels**.

Overall, these results should be of interest to policy-makers and managers, and our theoretical framework may provide a foundation upon which future researchers can answer XAI-related questions.

## 8. Suggestions for Future Work

In this paper, we model XAI through an economics lens. Specifically, explanations are considered an additional component of AI-based products that increase customer utility. Future researchers can look into other aspects of XAI such as number of explained features, explained R-squared, inferential vs. counterfactual explanations, etc. Moreover, recent research argues that in some situations, firms may not provide algorithmic transparency (Rubel, 2016) to avoid strategic manipulation by consumers who learn to game the AI decision-maker over repeated interactions (Wang et al., 2022). We believe further research is needed to investigate XAI in the presence of concerns about gaming by agents. Finally, we consciously exclude some factors such as adversarial attacks and model privacy issues in our model[a]. Future research can study the extent to which such factors may be affected by XAI in practical settings.

## 9. Funding and Competing Interests

The authors have no fundings or conflicts of interest to report.

[a] See Section 3.

# Appendix: Additional Results and Discussions

## A1. Local vs. Global Explanations

Broadly speaking, two classes of XAI methods have been proposed in the literature: (1) *global* explanations, which help us understand the entire model behavior, and (2) *local* explanations, which focus on a single prediction (Figure 11).

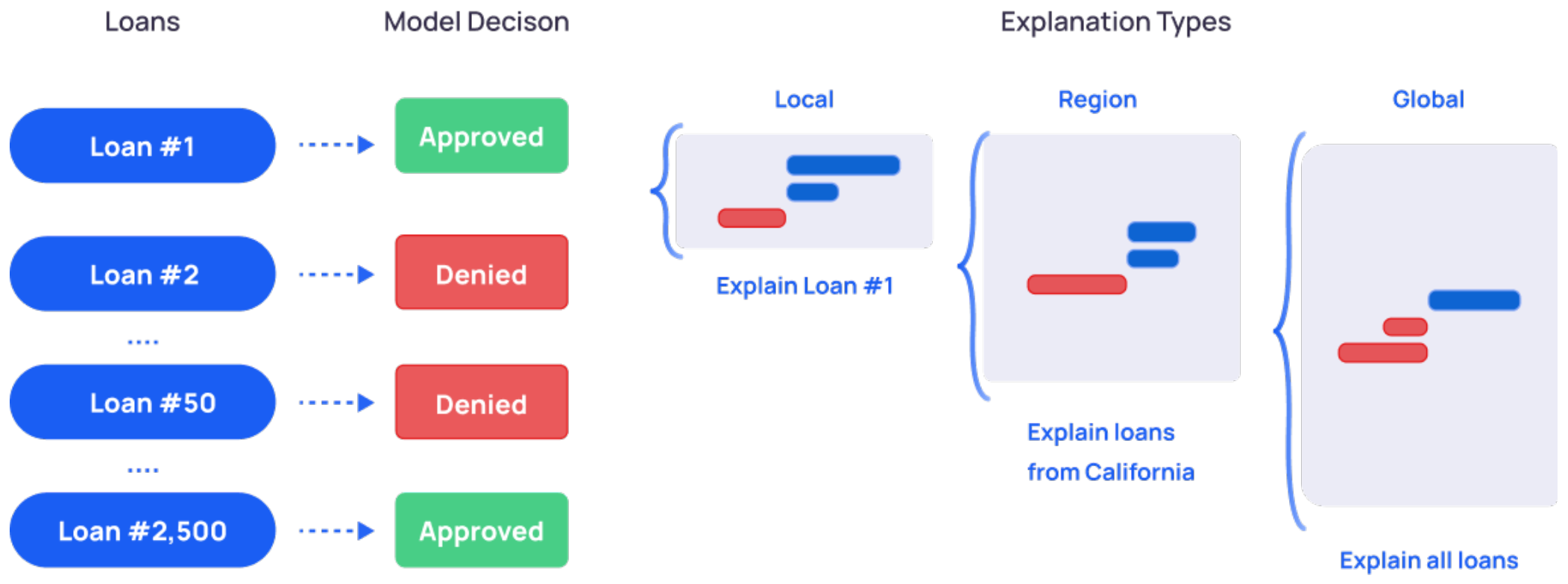

Figure 11. An example use of AI in loan applications. The AI either approves or rejects loan applicants, but does not readily tell them *why*. XAI not only provides an asnwer to this question, but it can also help analyze AI predictions in relation to the entire dataset or just a specific region to find anomalies and drifting data. Source: Fiddler Labs[a]

Global explanations are often useful for informing population-level decisions, such as climatic change or drug consumption trends (Yang et al., 2018) where an estimate of the global effect is more helpful than explaining every idiosyncratic possibility. On the other hand, local explanations aim to justify why the AI model made a specific decision or prediction for an instance. Most of the XAI scholarship focuses on this type of explanation, with methods such as LIME, LOCO, and SHAP to name a few (Lei et al., 2017; Lundberg and Lee, 2017; Ribeiro et al., 2016). The latter unifies local approaches using solid theoretical foundations of Shapley values from game theory (Roth, 1988; Shapley, 1951).

[a] https://www.fiddler.ai/explainable-ai

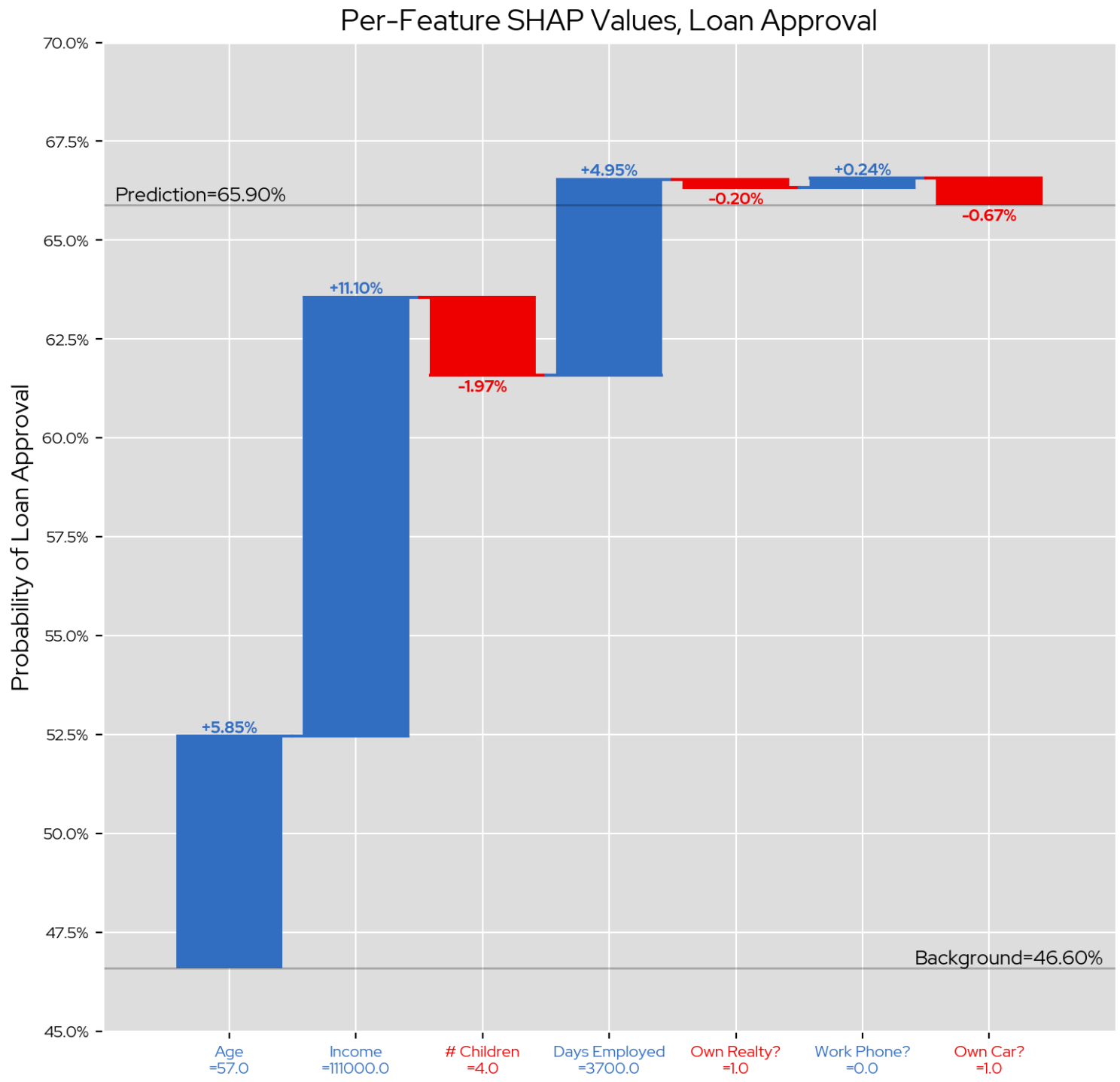

Figure 12. Shapley values for different features derived by the SHAP algorithm. As we can see, income and age have the highest positive effect on the probability of loan approval for this individual, whereas having 4 children has a negative effect. Source: KIE[a]

# A2. Equilibria, Indifference Lines, and Demands

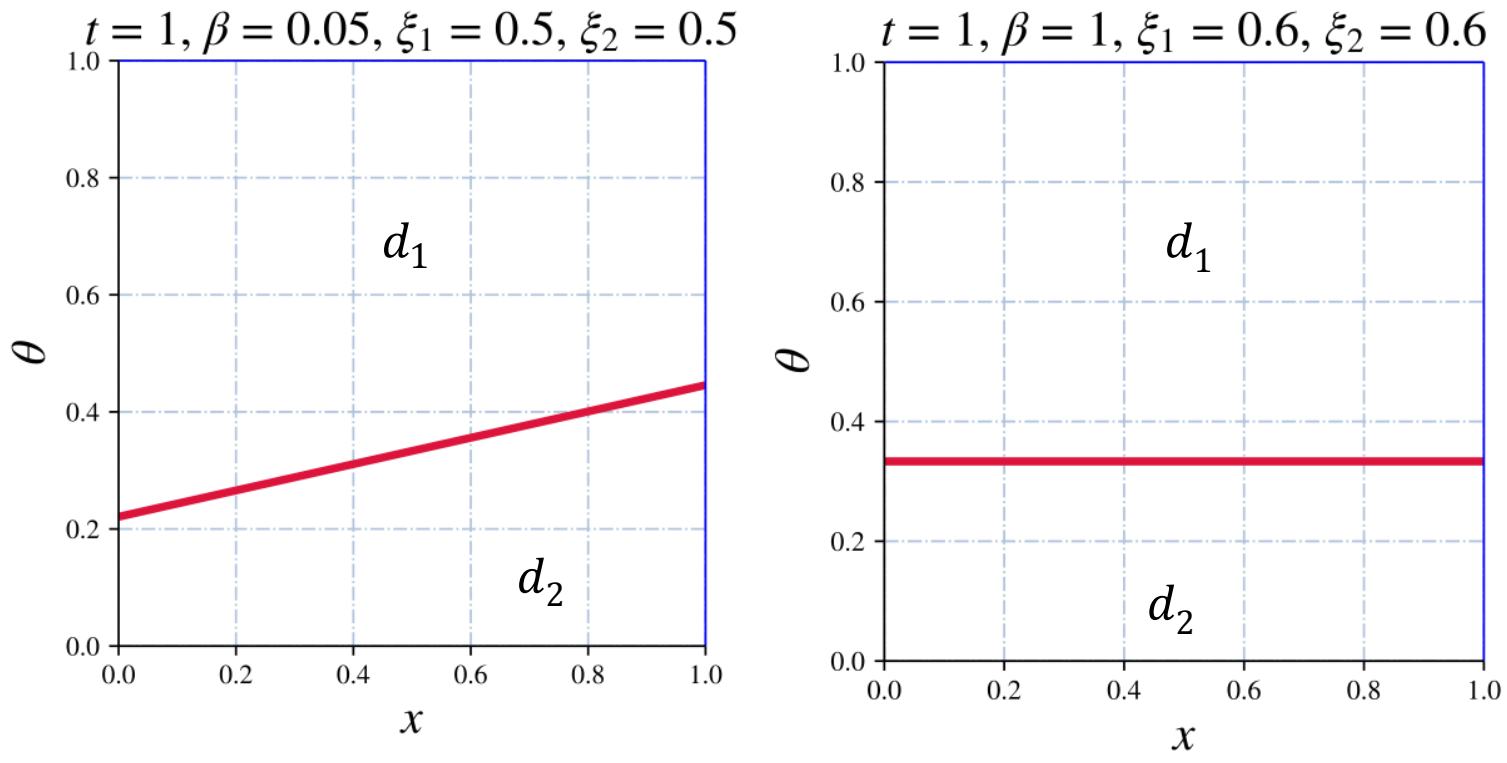

maxQ2 minQ2

Figure 13. Equilibria, indifference lines, and demands in market Q for some values of $\beta$ and $t$

[a] https://blog.kie.org/2021/11/trustyai-shap-overview-and-examples.html

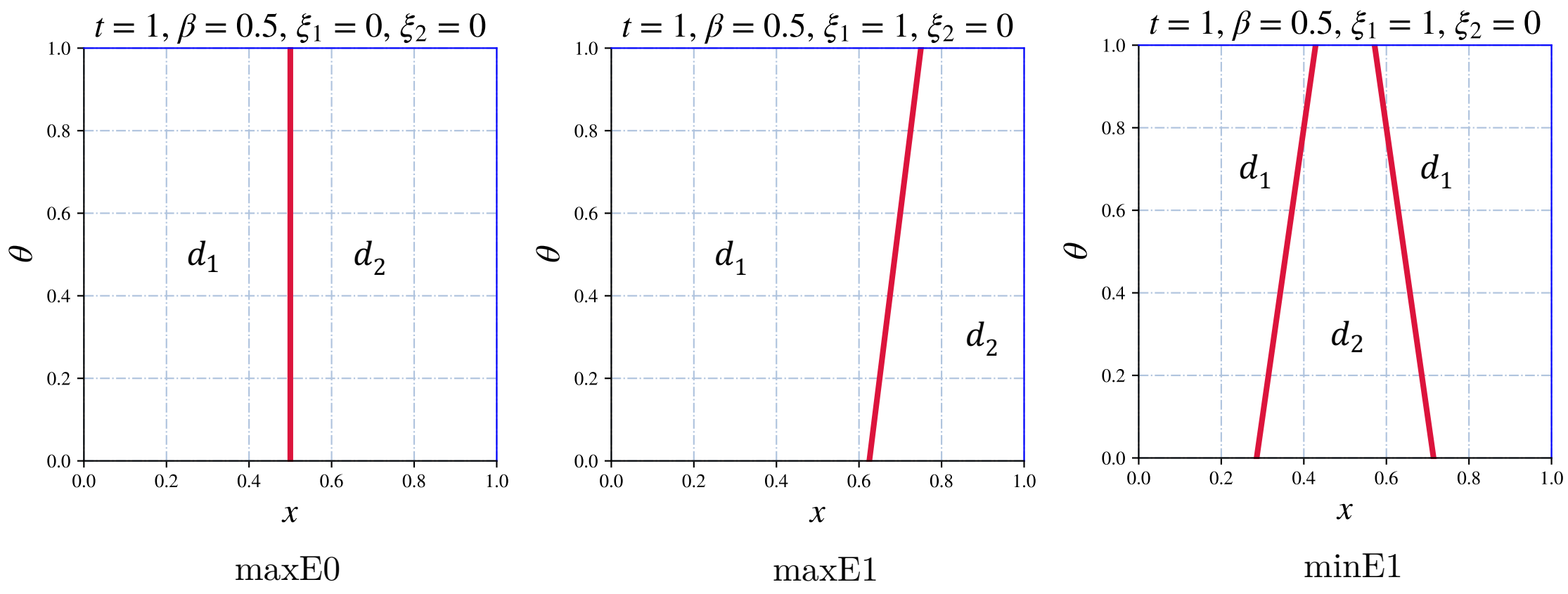

Figure 14. Equilibria, indifference lines, and demands in the E market[a] for $\beta = 0.5$ and $t = 1$

## A3. Proof of Lemma 1

Suppose that in addition to markets E and Q, a third market R exists, too. R can only be a combination of E and Q as follows:

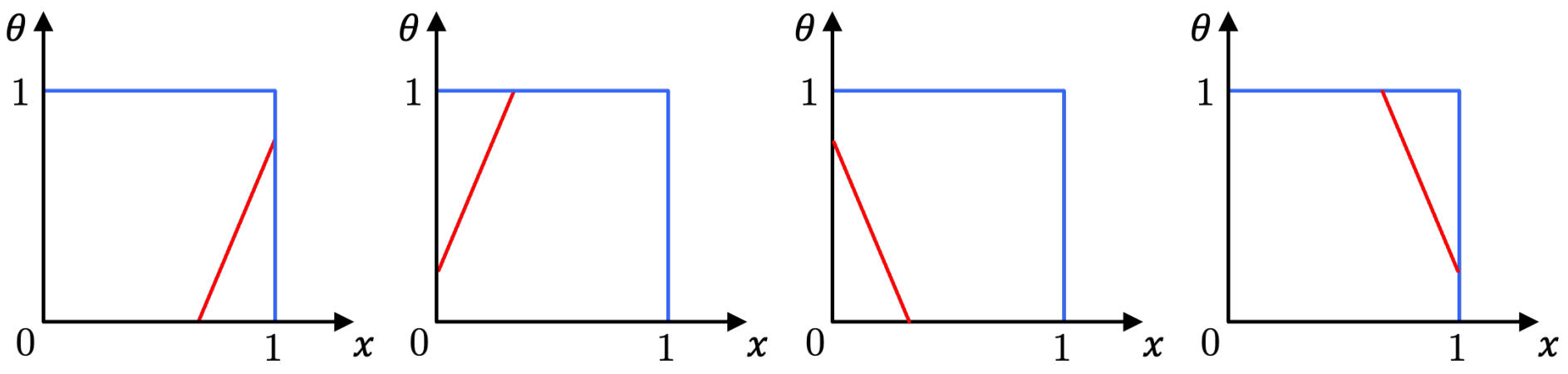

Figure 15. 4 possible combinations of markets E and Q

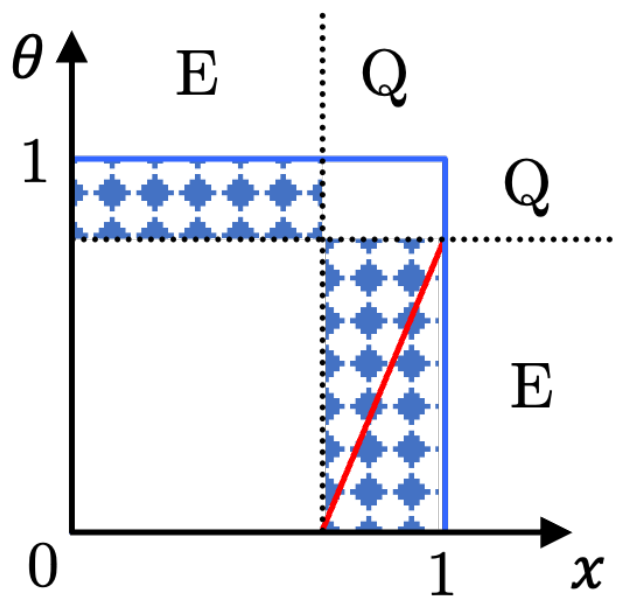

Now we show that none of these markets has a Nash equilibrium in our model. We run the proof for the first one; the rest are similar. Suppose that the first market in Figure 15 has a Nash equilibrium. Since preferences for quality and explanation are distributed uniformly throughout the unit square, one can break the hypothetical market R into submarkets E and Q[b]. The problem arises when we notice that two segments of R (shown by solid diamond grid) can be market E and Q at the same time, which is contradiction. In other words, no equilibrium exists for these regions and market R does not exist. ■

---

[a] Notice that due to the symmetry in minE1, the indifference line in $x \in [0, 0.5]$ will be reflected in $x \in [0.5, 1]$ as well, hence the two lines shown in Figure 14.

[b] Notice that breaking markets E and Q results in submarkets that are still E and Q, respectively.

## A4. Backward Induction to Solve the Games

First, we calculate the equilibrium prices $p_1^*$ and $p_2^*$ in stage 2 by solving $\partial\pi_1/\partial p_1 = 0$ and $\partial\pi_2/\partial p_2 = 0$, respectively. Next, we plug in these equilibrium prices in the profit functions and solve $\partial\pi_1/\partial q_1 = 0$ and $\partial\pi_2/\partial q_2 = 0$ to find the equilibrium quality levels $q_1^*$ and $q_2^*$, respectively. Finally, we compare the profit functions (if available; otherwise, we use their partial derivatives with respect to $\xi$) to determine which firm (if any) offers explanations. Doing this, we solve for the equilibrium decisions $(s_1^*, s_2^*)$ in stage 0.

## A5. Deriving the Equilibria

We derive all equilibria twice: once by hand and once by writing a computer code that verifies our results. For the latter, we use the Python programming language along with a Computer Algebra System (CAS) called Sympy[a], which is an open-source Python library for symbolic computation. Our Python code also enables us to run Monte-Carlo simulations when analytical solutions are not available (see Section 4.1.3).

## A6. Equilibrium Conditions

The conditions for equilibrium results mentioned in Table 2 are as follows:

Table 7. Equilibrium results mentioned in Table 2 along with their conditions

| Firms' Horizontal Differentiation | |
|---|---|
| max | min |
| maxE1<br>Condition: $\beta t > \frac{1}{6(2\xi^2-7\xi+6)}$ and $\underbrace{\xi > \xi_*}_{\beta t < 1/36(1-\xi)}$ | minE1<br>Condition: $\beta t > 2/9\xi$ |
| maxE0<br>Condition: $\beta t > \frac{1}{6(2\xi^2-7\xi+6)}$ and $\underbrace{\xi \leq \xi_*}_{\beta t \geq 1/36(1-\xi)}$ | minQ2<br>Condition: $\beta t < 4/9\xi$ |
| maxQ2<br>Condition: $\beta t < 4/9\xi$ | |

## A7. Firms' Equilibrium Profits in Market E

Profit functions in market E are shown below. As we can see, they are somewhat more complicated than profits in market Q (See Table 3).

[a] https://www.sympy.org/en/index.html

Table 8. Firms' equilibrium profits in market E

| Firms' Horizontal Differentiation | | $\pi_1$ | $\pi_2$ |
|---|---|---|---|
| max | If $\xi > \xi_*$: | If $\xi > \xi_*$: $\frac{(12\beta t\xi - 36\beta t + 1)^2(1 - 36\beta t\xi + 72\beta t)}{144\beta(18\beta t\xi - 36\beta t + 1)^2}$ | If $\xi \geq \xi_*$: $\frac{(24\beta t\xi - 36\beta t + 1)^2(1 - 36\beta t\xi + 72\beta t)}{144\beta(18\beta t\xi - 36\beta t + 1)^2}$ |
| max | If $\xi < \xi_*$: | If $\xi < \xi_*$: $\frac{t}{2} - \frac{1}{144\beta}$ | If $\xi < \xi_*$: $\frac{t}{2} - \frac{1}{144\beta}$ |
| min | | $\frac{(18\beta t\xi - 1)(12\beta t\xi - 1)^2}{144\beta(9\beta t\xi - 1)^2}$ | $\frac{(18\beta t\xi - 1)(6\beta t\xi - 1)^2}{144\beta(9\beta t\xi - 1)^2}$ |

# A8. Proof of Proposition 1

First, notice that both minE1 and minQ2 are possible when $2/9 < \beta t\xi < 4/9$. We start by proving that in this region, minE1 profits are always increasing in $\xi$ for both firms. Then, given that profits are constant in minQ2, we show that $\pi_1^{\text{minE1}}(\beta t\xi = 2/9) > \pi_1^{\text{minQ2}}(\beta t\xi = 2/9)$, indicating that firm 1 always earns more in minE1 than in minQ2. Conversely, we show that $\pi_2^{\text{minE1}}(\beta t\xi = 4/9) < \pi_2^{\text{minQ2}}(\beta t\xi = 4/9)$, implying that firm 2 always earns less in minE1 than in minQ2.

## A8.1. Part 1 of the Proof

Taking the derivative of $\pi_1$ w.r.t. $\xi$ we get:

$$\frac{\partial \pi_1}{\partial \xi} = \frac{t(12\psi - 1)(324\psi^2 - 81\psi + 4)}{24(9\psi - 1)^3}$$

where $\psi := \beta t\xi$. The terms $(12\psi - 1)$ and $(9\xi - 1)$ are positive for $\psi > 2/9$. Next, we find the roots of $324\psi^2 - 81\psi + 4 = 0$ and find that this expression in positive after its second root $\psi_2 = 0.81 < 2/9$. Therefore, the whole derivative is positive for the given region of $\beta t\xi$. Similar result holds for $\partial \pi_2/\partial \xi$:

$$\frac{\partial \pi_2}{\partial \xi} = \frac{t(6\psi - 1)(162\psi^2 - 27\psi + 2)}{24(9\psi - 1)^3}$$

which is always positive in the given range of $\beta t\xi$.

## A8.2. Part 2 of the Proof

Remember that $\pi_1^{\text{minE1}} = (12\psi - 1)^2(18\psi - 1)/(144\beta(9\psi - 1)^2)$. At $\psi = 2/9$ we get $\pi_1^{\text{minE1}}(\psi = 2/9) \approx 0.06/\beta$, which is already greater than $\pi_1^{\text{minQ2}} = 4/81\beta \approx 0.05/\beta$.

As with the second firm, $\pi_2^{\text{minE1}} = (6\psi - 1)^2(18\psi - 1)/(144\beta(9\psi - 1)^2)$. At $\psi = 4/9$ we have $\pi_2^{\text{minE1}} \approx 0.015/\beta$, which is still less than $\pi_2^{\text{minQ2}} = 2/81\beta \approx 0.025/\beta$. ■

## A9. Analytical Expressions of Total Welfares

Table 9. Analytical expression of total welfare at each equilibrium

| Equilibrium | Total Welfare ($W_t$) |
|---|---|
| maxE0 | $\dfrac{1-18\beta t}{36\beta}$ |
| maxE2 | $\dfrac{9\beta t(\xi-1)+1}{36\beta}$ |
| maxE1 | Too Long; available in our computer code. |
| maxQ2 | $\dfrac{27\beta t(\xi-1)+2}{54\beta}$ |
| minQ2 | $\dfrac{3.375e32\beta t\xi-3.375e32\beta t+5.0e31}{1.35e33\beta}$ where $xey$ means $x \times 10^y$. |
| minE1 | $\dfrac{1296\beta t^3\xi^3-1458\beta t^3\xi^2-90\beta t^2\xi^2+324\beta t^2\xi-21\beta t\xi-18\beta t+2}{72\beta(9\beta t\xi-1)^2}$ |

## A10. Analysis of Mandatory XAI Policies

The **red** region (!) in Figure 6 (left) is where firms earn negative profits in maxE2. This is found by setting $\pi_1 = \pi_2 = t(1-\xi)/2 - 1/144 < 0$ and solving for $\beta t$. The expression for the orange region can be found by comparing the smallest profit that firms can obtain in maxQ2 with its corresponding profit in maxE2. In this case, we need to compare $\pi_2^{\text{maxQ2}} = 2/81\beta$ with $\pi_2^{\text{maxE2}} = t(1-\xi)/2 - 1/144$ and solve for $\beta t$.

In the **purple** area, only market Q is possible. Since $W_t$ is positively correlated with $\xi$, the policy-maker chooses the maximum possible $\xi$ for any given $\beta t$. Therefore, for $\beta t < 1/6(2 \cdot 1^2 - 7 \cdot 1 + 6) = 1/6$ the policy-maker sets $\xi_{\text{policy}}^{\text{mand.}} = 1$ so that the market becomes Q and maximum $W_t$ is obtained. If the policy-maker would choose smaller $\xi$ (e.g., $\xi = 0.5$), then for some values of $\beta t$, firms would choose market E to earn higher profits and maximum total welfare would not be achieved.

Inside the **green** and **orange** regions, both markets E and Q are possible, but firms choose market Q in the orange region and E in the green region to maximize their profits. For $1/6 = \text{A} \leq \beta t < \text{B} = 4/9$, the policy-maker still sets $\xi_{\text{policy}}^{\text{mand.}} = 1$ so that the market remains Q and maximum $W_t$ is achieved. From point B to C, the policy-maker must choose $\xi_{\text{policy}}^{\text{mand.}} = 4/9\beta t - \varepsilon$ where $\varepsilon \to 0$, the reason being that $\xi = 4/9\beta t$ borders on the red region (!) in which $\pi_1, \pi_2 < 0$ and firms leave the market, resulting in $W_t = 0$. Thus, with the above $\xi_{\text{policy}}^{\text{mand.}}$, the market remains Q from B to C and firms earn positive profits.

In the **blue** region, the policy-maker chooses $\xi_{\text{policy}}^{\text{mand.}} = 1 - (1/72)/\beta t$ from point C to D onward and the market is now E. Under this policy, from point C onward, firms earn zero profit and they cannot form a Q market from C to D due to the choice of $\xi^{\text{mand.}}_{\text{policy}}$ by the policy-maker.

Table 10. 5 regions on the $\beta t$–$\xi$ plane under mandatory XAI and firms use different XAI methods

| Region(s) | Which Market is Possible? | Relation Between $\beta t$ and $\xi$ |
|---|---|---|
| Blue and Red | only E | In Red and Blue: $\beta t > 4/9\xi$ |
| | | In Red: $\beta t < (1/72)/(1-\xi)$ |
| Green and Orange | E and Q | In Green and Orange: $1/6(2\xi^2 - 7\xi + 6) < \beta t < 4/9\xi$ |
| | | In Orange: $\beta t < 0.063/(1-\xi)$ |
| Purple | only Q | $\beta t < 1/6(2\xi^2 - 7\xi + 6)$ |

# A11. Number of Firms that Offer XAI w.r.t. $\xi$

Figure 9 is obtained from Figure 5 as follows: When firms use the same XAI method, for small $\beta t$ both firms offer XAI regardless of $\xi$—the green region in Figure 5, right. Therefore, the number of firms that offer XAI is constant in this case. Now let us increase $\beta t$. If $\xi$ is small, then both firms offer XAI, but if it is large, then only one firm offers XAI. In the gray region, there is no equilibrium, and no firm offers XAI.

On the other hand, when firms use different XAI methods, Figure 5, left, tells us that with small $\beta t$, both firms offer XAI for all values of $\xi$—the green region in the figure. As $\beta t$ increases, there is a region where increasing $\xi$ results in switching from "No Nash Equilibrium" (gray region) to both firms offering XAI. After that, increasing $\beta t$ results in a situation where no firm offers XAI (the blue and gray regions in Figure 7) for all $\xi$. Finally, as $\beta t$ keeps increasing, we see that at most one firm offers XAI—the yellow region.

# A12. Analysis of XAI Fairness

## A12.1. Proof of the Condition for Vertical Fairness

Vertical fairness requires that for all $x_j \in [0, 1]$, customers on the vertical line $x = x_j$ receive the same XAI level (See Figure 16). If this line completely falls inside a firm's market, then vertical fairness holds. But in both explanation- and quality-dominated markets, there exist(s) $x_j \in \mathcal{X}, |\mathcal{X}| \geq 1$ where $x = x_j$ intersects with the indifference line. Thus, some customers on this line fall in firm 1's market and the rest

fall in firm 2'. Therefore, for vertical fairness to hold, both firms must offer the same level of XAI. ■

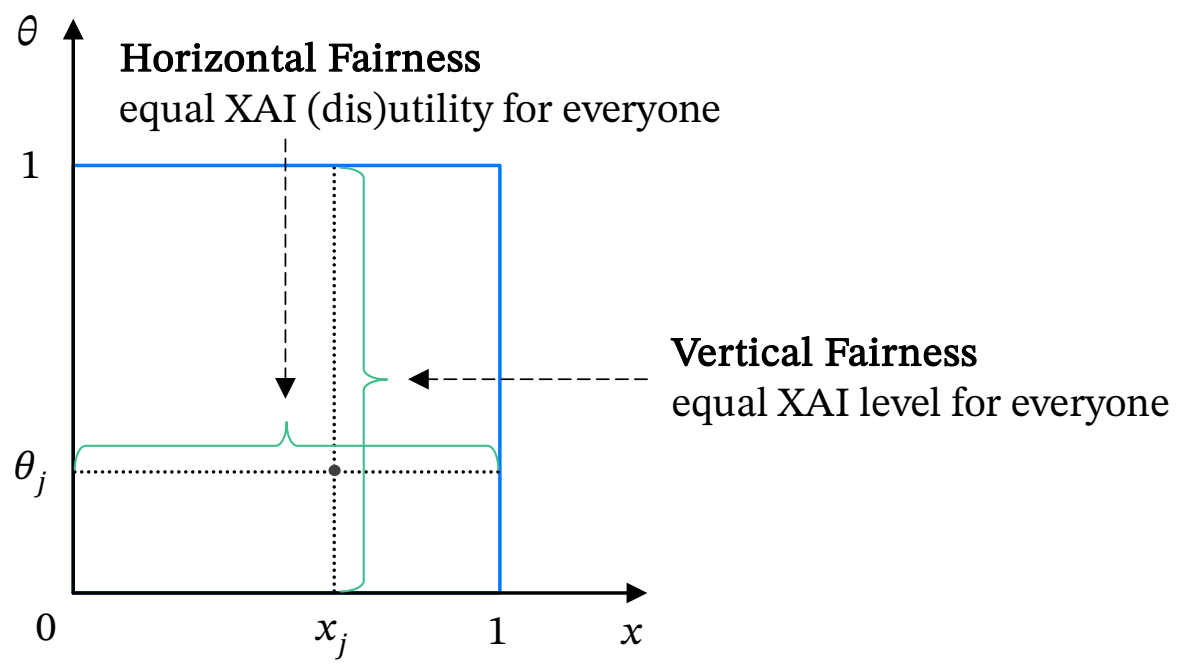

Figure 16. Vertical and Horizontal XAI fairness

## A12.2. Two Types of XAI Fairness in Each Equilibrium

All possible scenarios under optional and mandatory XAI policies are listed below:

Table 11. All possible scenarios in our model under optional and mandatory XAI policies. 'O' and 'M' stand for 'optional' and 'mandatory', respectively. Vertical and horizontal fairness are denoted by 'V' and 'H', respectively.

| | Scenario | maxQ2 | minQ2 | maxE0 | maxE1 | minE1 | maxE2 |
|---|---|---|---|---|---|---|---|
| | Equilibrium Under ... | O, M | O, M | O | O | O | M |
| XAI Fairness | V | ✅ | ✅ | ✅ | ❌ | ❌ | ✅ |
| XAI Fairness | H | M:<br>✅ if C1<br>❌ if C2<br>O: ✅ | M:<br>✅ if C1<br>❌ o.w.<br>O:<br>❌ if C3<br>✅ o.w. | ❌ | ❌ | ❌ | M: ❌ |

Conditions: C1 $\longleftrightarrow \beta t < 4/9$, C2 $\longleftrightarrow \beta t \in 4/9 \times (1, 1/0.97)$, C3 $\longleftrightarrow \beta t \in (0.225, 0.445)$

## A12.3. Tradeoffs

### A12.3.1. Vertical fairness vs. Maximum Total Welfare

Under optional XAI, Figure 5 illustrates that there exists a tradeoff between maximizing $W_t$ and achieving vertical fairness, regardless of firms' horizontal differentiation ('min' or 'max'). In 'max', notice that the policy-maker chooses $\xi = 1$ (the yellow region) to maximize $W_t$, whereas he would need to set $\xi < 1 - 1/36\beta t$ (bold line —) to obtain vertical fairness through maxE0 (the blue region). Similarly, in 'min' the policy-maker could either choose $\xi = 1$ (the yellow region) and maximize $W_t$, or set $\xi < 2/9\beta t$ (bold line —) to support vertical fairness through minQ2 (the green region).

There is no tradeoff between maximizing $W_t$ and attaining vertical fairness under mandatory XAI, because mandatory XAI forces both firms to offer XAI at the same level, reaching maximum welfare and vertical fairness simultaneously.

### A12.3.2. Horizontal fairness vs. Maximum Total Welfare

Under optional XAI, there is no tradeoff between maximizing $W_t$ and achieving horizontal fairness. Figure 5 shows that when firms use different XAI methods, the only time horizontal fairness can happen is in scenario maxQ2 where $\beta t < 0.165$. In this scenario, the policy-maker who maximizes $W_t$ sets $\xi_{\text{policy}}^{\text{mand.}} = 1$, so horizontal fairness is achieved as well. Similar story holds for minQ2 ($\beta t < 0.225$) when firms use the same XAI method. Notice that choosing minQ2 over minE1 for $\beta t > 0.445$ does not result in horizontal fairness because $\xi \neq 1$.

Similarly, there is no tradeoff between horizontal fairness and maximizing $W_t$ under mandatory XAI. Thinking about the minQ2 equilibrium and the fact that the policy-maker sets $\xi_{\text{policy}}^{\text{mand.}} = \min\{4/9\beta t\,, 1\}$, one might conclude that $W_t$ is maximized at the cost of horizontal fairness if $\beta t > 4/9$. But setting $\xi_{\text{policy}}^{\text{mand.}} = 1$ will only maximize $W_t$ when $\beta t < 4/9$ (and it happens that horizontal fairness is achieved in this case). For $\beta t > 4/9$, having $\xi_{\text{policy}}^{\text{mand.}} = 1$ is simply not possible because it violates the condition of minQ2. Moreover, notice that when firms use different XAI methods, the policy-maker sets $\xi_{\text{policy}}^{\text{mand.}} < 1$ in the red region (!) to allow for a Nash equilibrium and maximize total welfare, thus there is no tradeoff between maximizing $W_t$ and attaining horizontal fairness as $\xi_{\text{policy}}^{\text{mand.}} = 1$ is simply not possible.

# A13. Unregulated XAI Level

## A13.1. maxE (Full Model)

Firms play the same game as in the exogenous case, except that their decisions in stage-0 are not about offering/not offering XAI, but rather about the level of explanations that they offer. We can solve this by plugging $p^*$ and $q^*$ in the profit functions and solving $\partial\pi_1/\partial\xi_1 = 0$ and $\partial\pi_2/\partial\xi_2 = 0$ to find the stage-0 equilibrium levels of explanations $\xi_1^*$ and $\xi_2^*$, respectively. The equilibrium results are:

$$p_1^* = \frac{-t(\xi_1 + \xi_2 - 2)(12\beta t(\xi_1 + 2\xi_2 - 3) + 1)}{2(18\beta t(\xi_1 + \xi_2 - 2) + 1)};$$

$$p_2^* = \frac{-t(\xi_1 + \xi_2 - 2)(12\beta t(\xi_2 + 2\xi_1 - 3) + 1)}{2(18\beta t(\xi_1 + \xi_2 - 2) + 1)};$$

$$
\begin{aligned}
q_1^* &= \frac{12\beta t(\xi_1 + 2\xi_2 - 3) + 1}{12\beta(18\beta t(\xi_1 + \xi_2 - 2) + 1)};\\
q_2^* &= \frac{12\beta t(\xi_2 + 2\xi_1 - 3) + 1}{12\beta(18\beta t(\xi_1 + \xi_2 - 2) + 1)};\\
\xi_1^* &= 1 - \frac{1}{36\beta t};\\
\xi_2^* &= 1 - \frac{1}{36\beta t};\\
\Delta p &= \frac{6\beta t^2(\xi_1 - \xi_2)(\xi_1 + \xi_2 - 2)}{18\beta t(\xi_1 + \xi_2 - 2) + 1};\\
\Delta q &= \frac{-t(\xi_1 - \xi_2)}{18\beta t(\xi_1 + \xi_2 - 2) + 1};\\
\pi_1 &= -\frac{(12t\beta\xi_1 + 24t\beta\xi_2 - 36t\beta + 1)^2(36t\beta\xi_1 + 36t\beta\xi_2 - 72t\beta + 1)}{144\beta(18t\beta\xi_1 + 18t\beta\xi_2 - 36t\beta + 1)^2};\\
\pi_2 &= -\frac{(24t\beta\xi_1 + 12t\beta\xi_2 - 36t\beta + 1)^2(36t\beta\xi_1 + 36t\beta\xi_2 - 72t\beta + 1)}{144\beta(18t\beta\xi_1 + 18t\beta\xi_2 - 36t\beta + 1)^2}.
\end{aligned}
$$

Notice that this is a symmetric equilibrium in $p^*$, $q^*$, and $\xi^*$.

## A13.2. minE (Full Model)

If we solve for equilibrium XAI levels in this case, we obtain the following system of equations:

$$
\begin{aligned}
\xi_1 &= \frac{12\beta t\xi_2 + 1}{12\beta t};\\
\xi_2 &= \frac{6\beta t\xi_1 - 1}{6\beta t} \Rightarrow \xi_1^* = \xi_2^* + \frac{1}{6\beta t}
\end{aligned}
$$

which is overdetermined and has no solution. Therefore, sustaining market E while using the same XAI method is not possible.

## A13.3. Simplified Model

The game still has three stages:

- Stage 0: firms choose $\xi_1^*, \xi_2^* \in [0, 1]$;
- Stage 1: firms choose $q_1^*, q_2^* \in \{0, q\}$;
- Stage 2: firms choose $p_1^*, p_2^* \in \{0, p\}$.

We only examine equilibria where both firms enter the market, i.e., they choose $p_1^* = p_2^* = p$ in stage 2. Again, backward induction is used to solve for equilibrium qualities and XAI levels. In each stage, we find firm $i$'s best response to the other firm's action.

### A13.3.1.Firms Undifferentiated ('min') in XAI Methods

Only the explanation-dominated market is viable. We identify an equilibrium where unregulated firms offer partial XAI ($\xi < 1$) in equilibrium:

$$p_1^* = p_2^* = p; \quad q_1^* = q_2^* = 0; \quad \xi_1^* = \xi_2^* = 1 - \eta/2$$

where $\eta \stackrel{\text{def}}{=} 1/2\beta t \times p/q$. This equilibrium corresponds to maxE2 in the full model and is sustained if $p/\beta q^2 \in [1, 2]$ and $\beta q \leq 1/4$. The total welfare $W_t$ and total consumer utility in unregulated XAI are better than regulated XAI when $\eta \in [1.33, 2]$, worse when $\eta \in [1, 1.33]$, and the same otherwise. To get a sense of this equilibrium, see firm payoffs in Table 12.

Table 12. The game between firms in stage 2 of the simplified model. In each cell, the top row is $\pi_1$ and the bottom row $\pi_2$.

| | | Firm 1's quality choice ($q_1$): 0 | Firm 1's quality choice ($q_1$): $q$ |
|---|---|---|---|
| Firm 2's quality choice ($q_2$) | 0 | $\xi_r p$<br>$(1-\xi_r)p$ | $(\xi_r + q/2t\bar{\xi})p - \beta q^2$<br>$\left((1-\xi_r) - q/2t\bar{\xi}\right)p$ |
| | $q$ | $(\xi_r - q/2t\bar{\xi})p$<br>$\left((1-\xi_r) + q/2t\bar{\xi}\right)p - \beta q^2$ | $\xi_r p - \beta q^2$<br>$(1-\xi_r)p - \beta q^2$ |

where $\bar{\xi} \stackrel{\text{def}}{=} 1 - \xi_1 + 1 - \xi_2$ and $\xi_r \stackrel{\text{def}}{=} (1-\xi_2)/\bar{\xi}$. With symmetric XAI ($\xi_1 = \xi_2, \xi_r = 0.5$) and zero quality ($q_1 = q_2 = 0$), each firm earns a payoff of $p/2$. As firm 1 (firm 2) increases XAI $\xi_r$ increases (decreases) and so does firm 1's (firm 2's) market share. If firm 1 increases quality ($q_1 = q$) they get additional market share $q/2t\bar{\xi}$ but pay cost $\beta q^2$.

### A13.3.2. Firms Undifferentiated ('min') in XAI Methods

Similar setup as before. Only quality dominated Q-market is viable. We identify an equilibrium where unregulated firms offer less than full XAI in equilibrium:

$$p_1^* = p_2^* = p; \quad (q_1^*, q_2^*) = (q, 0); \quad (\xi_1^*, \xi_2^*) = (1-\delta, 1)$$

where $\delta \stackrel{\text{def}}{=} 4q/t \times (1 - \beta q^2/p)$. This equilibrium corresponds to minQ2 in the full model and holds when $p/\beta q^2 \in [1, 2]$. Since mandatory XAI is not feasible, we only compare unregulated and optional XAI. Interestingly, total welfare, total consumer utility, and average XAI level are all higher in unregulated XAI than in optional XAI.

## Web Links

1 https://www.gartner.com/en/information-technology/insights/top-technology-trends

2 https://www.idc.com/getdoc.jsp?containerId=IDC_P33198

3 https://www.newyorker.com/magazine/2017/04/03/ai-versus-md

4 https://www.bloomberg.com/graphics/2016-amazon-same-day/

5 https://www.inverse.com/article/39483-apple-refund-china

6 https://futurism.com/ai-bias-black-box

7 https://www.darpa.mil/program/explainable-artificial-intelligence

8 https://ec.europa.eu/futurium/en/ai-alliance-consultation.1.html

9 https://tinyurl.com/theguardian-AI-watchdog

10 https://www2.deloitte.com/us/en/insights/industry/financial-services/explainable-ai-in-banking.html

11 https://research.aimultiple.com/xai/

12 https://tinyurl.com/theguardian-transparent-ai

13 http://www.ftc.gov/legal-library/browse/statutes/fair-credit-reporting-act

14 https://insight.equifax.com/visualizing-xai-in-credit-risk/

15 https://www.ntsb.gov/news/events/Pages/2020-HWY18FH011-BMG.aspx

16 https://www.compact.nl/en/articles/autonomous-compliance/

17 https://www.wired.com/story/europes-new-privacy-law-will-change-the-web-and-more/

18 https://tinyurl.com/vox-GDPR-data-protection

19 https://tinyurl.com/cnet-GDPR-fb-google

20 https://tinyurl.com/forbes-GDPR-benefits

21 https://tinyurl.com/vox-GDPR-data-protection

22 https://tinyurl.com/cnbc-fb-call-for-GDPR

23 https://tinyurl.com/comply-or-explain

24 https://www.wipo.int/news/en/wipolex/2016/article_0014.html

25 https://tinyurl.com/GDPR-harmful-for-AI

26 https://tinyurl.com/cnbc-Intel-AI-infancy